\documentclass[10pt]{article} %
\usepackage[preprint]{tmlr}

\usepackage{amsmath,amsfonts,bm}

\def\eqref#1{equation~\ref{#1}}

\def\1{\bm{1}}

\DeclareMathAlphabet{\mathsfit}{\encodingdefault}{\sfdefault}{m}{sl}
\SetMathAlphabet{\mathsfit}{bold}{\encodingdefault}{\sfdefault}{bx}{n}

\usepackage{graphicx} 
\usepackage{subcaption} %
\usepackage{pgfplots}   %
\usepgfplotslibrary{groupplots}
\pgfplotsset{compat=1.18}
\usepackage{hyperref}
\usepackage{url}
\usepackage{booktabs}
\usepackage{multirow}
\usepackage{makecell}   %
\usepackage{colortbl}   %
\usepackage{algpseudocode} %
\usepackage{pifont}
\newcommand{\xmark}{\ding{55}}
\newcommand{\cmark}{\ding{51}}
\usepackage[table]{xcolor}
\usepackage[normalem]{ulem}
\usepackage{xspace}
\usepackage{array}
\usepackage{pifont}     %
\usepackage[noabbrev,nameinlink]{cleveref} %
\usepackage{algpseudocode}
\usepackage{algorithm}
\usepackage{wrapfig}
\usepackage{svg}
\algrenewcommand\algorithmicrequire{\textbf{Input:}}
\algrenewcommand\algorithmicensure{\textbf{Output:}}

\crefname{section}{sec.}{secs.}
\Crefname{section}{Sec.}{Secs.}

\newcommand{\tablestyle}[2]{\setlength{\tabcolsep}{#1}\renewcommand{\arraystretch}{#2}\centering\footnotesize}
\newcolumntype{H}{>{\setbox0=\hbox\bgroup}c<{\egroup}@{}}
\definecolor{ResultGreen}{rgb}{0.0, 0.5, 0.0}

\newcommand{\ours}{UnSAM\xspace}

\newcommand{\mymethod}{Diffuse2Seg\xspace}
\newcommand{\mymethodplus}{Diffuse2Seg-Semi\xspace}
\newcommand{\cutler}{CutLER\xspace}
\newcommand{\eg}{\textit{e.g.,}\xspace}

\usepackage{svg}
\svgpath{{../imgs/}} %

\definecolor{abBlue}{HTML}{0072B2}
\definecolor{abOrange}{HTML}{D55E00}
\definecolor{abTeal}{HTML}{009E73}
\definecolor{abPurple}{HTML}{CC79A7}
\definecolor{abGrey}{HTML}{555555}

\pgfplotsset{
  ablation panel/.style={
    axis line style={black!55,line width=.4pt},
    tick style={black!55,line width=.4pt},
    grid style={black!12,line width=.3pt},
    ymajorgrids=true,
    label style={font=\scriptsize},
    tick label style={font=\scriptsize},
    title style={font=\scriptsize,yshift=-2pt},
    legend cell align=left,
    legend style={
      font=\scriptsize,draw=black!30,fill=white,fill opacity=.85,
      text opacity=1,inner sep=1.5pt,row sep=-1pt,
    },
    unbounded coords=discard,          %
    clip marker paths=true,
  },
  ablation curve/.style={
    abBlue,line width=.8pt,mark=*,mark size=1.4pt,
    mark options={solid,fill=abBlue,draw=abBlue},
  },
  ablation selected/.style={
    only marks,mark=star,mark size=3.4pt,
    abOrange,mark options={line width=.7pt},
  },
  ablation best/.style={
    only marks,mark=o,mark size=2.8pt,
    abTeal,mark options={line width=.7pt},
  },
  ablation curve map/.style={
    abPurple,line width=.8pt,densely dotted,mark=triangle*,mark size=1.7pt,
    mark options={solid,fill=abPurple,draw=abPurple},
  },
  ablation curve alt/.style={
    abOrange,line width=.8pt,dashed,mark=square*,mark size=1.4pt,
    mark options={solid,fill=abOrange,draw=abOrange},
  },
  ablation curve third/.style={
    abPurple,line width=.8pt,dotted,mark=triangle*,mark size=1.8pt,
    mark options={solid,fill=abPurple,draw=abPurple},
  },
}

\usepackage{algorithmicx}

\newif\ifdraft
\usepackage{color}

\drafttrue

\ifdraft
\usepackage{cancel}
\usepackage{soul}
\fi

\usepackage{changepage}
\usepackage{microtype}

\title{\mymethod: Diffusion Models Can Segment Anything\\ Without Supervision}

\author{\name Christoph Hümmer \email christoph.huemmer@cariad.technology \\
\addr Institute of Mathematics, Technical University of Berlin \\
CARIAD SE, Volkswagen Group
      \AND
      \name Joachim Sicking \email joachim.sicking@cariad.technology \\
      \addr CARIAD SE, Volkswagen Group
      \AND
      \name Fabian Hüger \email fabian.hueger@cariad.technology \\
      \addr CARIAD SE, Volkswagen Group
      \AND
      \name Hanno Gottschalk \email gottschalk@math.tu-berlin.de \\
      \addr Institute of Mathematics, Technical University of Berlin}

\def\month{MM}  %
\def\year{YYYY} %
\def\openreview{\url{https://openreview.net/forum?id=XXXX}} %
\begin{document}
\maketitle
\begin{abstract}
Open-world entity segmentation aims to predict masks for arbitrary objects across domains and at multiple granularities, from parts to whole objects. In this setting, SAM sets a strong standard: trained on SA-1B, comprising 11M images and over 1B carefully annotated masks, it achieves remarkable zero-shot performance. Collecting such labels is expensive and time-consuming, however, which limits how far this recipe can scale. Text-to-image diffusion models offer a way around this. Their intermediate features transfer well across perception tasks, and since object structure emerges as the model denoises a noise sample into an image conditioned on a text prompt, that structure is already encoded in these representations. They can therefore be exploited for open-world entity segmentation without retraining or supervision. Building on this observation, we introduce \textit{\mymethod}, which repurposes generative diffusion models for automatic mask generation by propagating a grid of point prompts through their self-attention representations in an edge-preserving manner. \mymethod produces multi-granular instance masks and outperforms prior state-of-the-art label generators by 4.3-7.1 p.p.\ in $\mathrm{AR}_{1000}$ across five domains. Training an instance segmentation model on these generated masks advances detector-free open-world segmentation by 7.4 and 7.7 p.p.\ on ``things'' and ``stuff+things'' datasets and surpasses the detector-based UnSAM on ``stuff+things'' by 2.1 p.p.\ in $\mathrm{AR}_{1000}$. Finally, we show that a model trained on \mymethod labels provides a strong initialization for semi-supervised learning, outperforming its fully supervised counterpart with already 5k labeled images.

\end{abstract}
\section{Introduction}

\begin{figure}[!t]
	\centering
	\includegraphics[width=\linewidth]{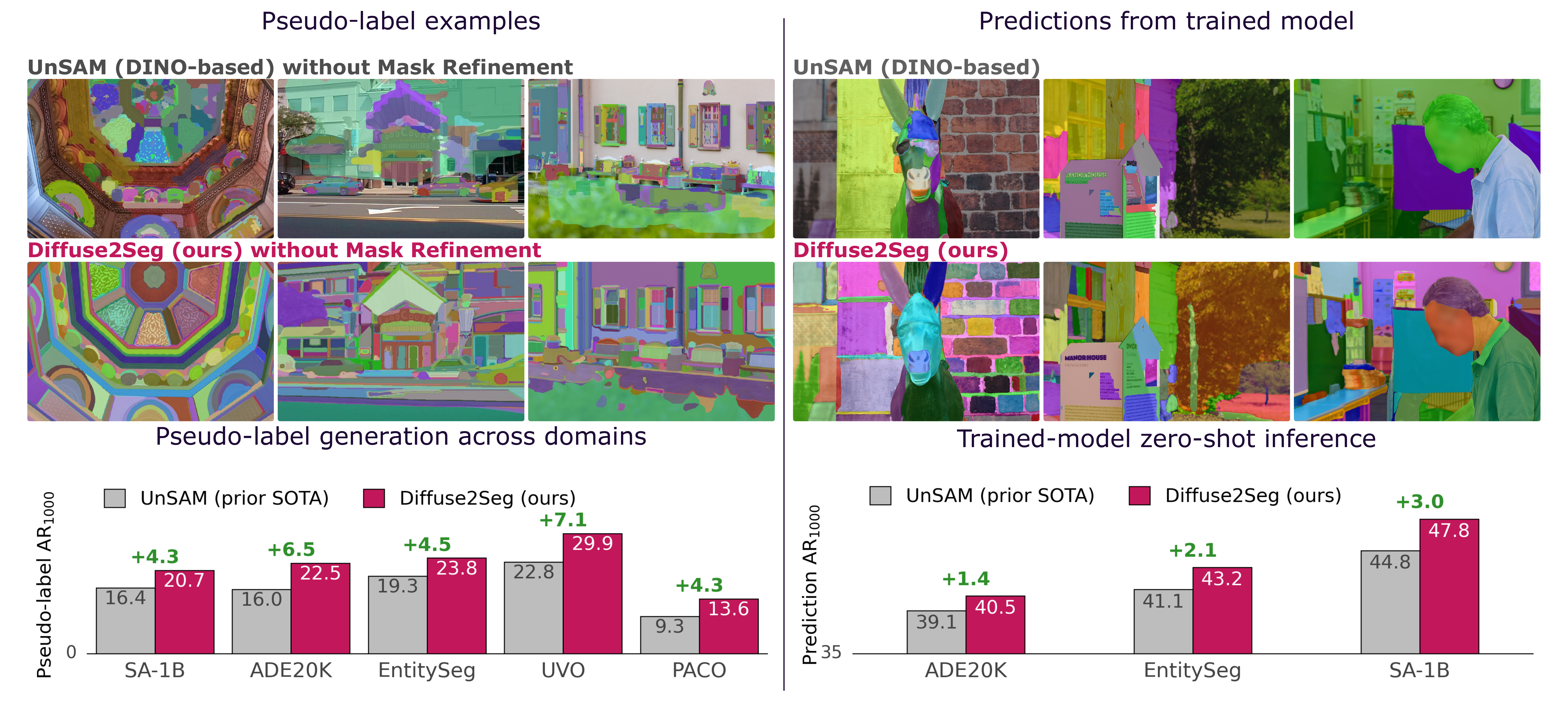}
	 \caption{\textbf{\mymethod: Multi-granular pseudo-labels from diffusion-based self-attention. } Our method produces smooth masks with dense coverage via diffusion self-attention, outperforming prior ``things''- biased DINO-based state-of-the-art UnSAM in pseudo-label average recall (AR). Training models on our diffusion-based labels improves prior state-of-the-art performance, particularly on dense segmentation datasets covering both ``stuff'' and ``things''. 
    }
	\label{fig:qualitative-coverage}
\end{figure}

Segmenting \emph{anything} means partitioning an image into all of its coherent visual entities, countable ``thing'' objects such as persons and cars as well as amorphous ``stuff'' regions such as sky and road, and doing so without being restricted to a fixed set of class labels \citep{qi2022open}. Such a segmentation generalist enables computer vision systems to operate in the open world without retraining and is therefore highly relevant in domains that constantly need to adapt, such as robotics or medical imaging. SAM \citep{kirillov2023segany} brought this concept to life: a foundation model trained over multiple rounds on the large SA-1B dataset of class-agnostic, high-quality masks covering the full object hierarchy from part to object, with strong zero-shot capabilities. However, densely annotating such a dataset needed to train open-world entity segmentation foundation models is particularly challenging and time-consuming, limiting scalability. This raises the demand to generate multi-granularity entity labels including things, stuff and parts with no manual supervision at all. %

One research direction tackling this issue lies in unsupervised instance mask generation from pre-trained foundation models, mainly using representations from self-supervised learning (SSL)-pretrained ViTs \citep{wang2023cut, arica2024cuvler, wang2024segment, Sick_2025_ICCV, yang2025unmore}. Nevertheless, these SSL-based approaches are mostly designed for foreground instance segmentation, and only a few works explore truly dense entity annotation like SOHES \citep{cao2024sohes} and UnSAM \citep{wang2024segment}. In parallel, text-to-image diffusion models have emerged as universal representation learners \citep{gabeur2026image, wang2026genception}, as generative synthesis naturally forces the network to model global scene composition, contextual relationships, and fine-grained spatial layout. That means if a diffusion models generates sky or a car from an input prompt, these entities should be extractable from the internal representations. A large body of work has already shown that the features of these models can be leveraged for tasks like unsupervised semantic segmentation at different granularities \citep{tian2024diffuse, couairon2024diffcut,namekata2024emerdiff, kim2025seg4diff, zhang2026falcon, nakhli2026unlocking}, interactive segmentation \citep{M2N2, karmann2025m2n2v2} or enhancing SSL-based instance segmentors \citep{jo2026trace}. 

Pursuing this path, we propose \textit{\mymethod}, 
an unsupervised entity-segmentation method that repurposes pretrained diffusion models to generate class-agnostic multi-granular instance labels. 
\mymethod is the first method to do so without retraining the diffusion model.
Our key observation is that we can
exploit emerging capabilities of a generative diffusion model for class-agnostic multi-granular instance label generation: we apply non-linear p-Laplacian-based propagation 
to a grid of dense point prompts using an efficiently extracted self-attention matrix to obtain soft object masks. These are then merged into multiple granularity segmentation maps and instances are isolated. Finally, we aggregate and remove duplicates from the resulting mask candidates, producing instance masks at various granularities at the original image resolution. 
Moreover, we show that the resulting labels are a valuable supervision signal by distilling them into a lightweight segmentation model, which yields 
strong zero-shot generalization. In addition, we demonstrate that our unsupervised model can be leveraged for data-efficient semi-supervised learning through simple decoder-only fine-tuning when limited ground-truth data is available.

\newpage
In summary, we make three contributions:

\begin{adjustwidth}{0.5cm}{0.5cm}
\textbf{(i) Training-free multi-granularity instance segmentation} We show that a frozen diffusion model can act as an automatic mask generator: by propagating a grid of point prompts through diffusion model self-attention, we recover dense instance masks at multiple granularities, entirely without supervision. Our generated labels surpass the previous unsupervised SOTA, UnSAM, by \textbf{4.3\,-\,7.1} p.p.\ in $\text{AR}_{1000}$ across a diverse set of five domains. 

\textbf{(ii) State-of-the-art pseudo-label training} We propose training a lightweight instance segmentation model solely on our diffusion-based pseudo-labels outperforming the strongest detector-free baseline, SOHES, by \textbf{+7.7/+7.4/+10.0} p.p.\ AR on stuff+things/things/parts datasets, at half of the training image budget and using a much smaller backbone and improve the prior state-of-the-art technique UnSAM in `stuff+things' by \textbf{+2.1} p.p.\ in AR in a detector-augmented setting.

\textbf{(iii) Semi-supervised \mymethod-Semi} We apply a simple supervised fine-tuning to our unsupervised model and the prior state-of-the-art and compare them to a strong reference trained on 100k SA-1B ground-truth labels on our extensive zero-shot benchmark. Our fine-tuned semi-supervised model recovers 93.5 \% of a fully supervised reference at only 100 fine-tuning images and even surpasses it by 1.4 p.p.\ at 10k fine-tuning images, maintaining a consistent advantage compared to the prior state-of-the-art UnSAM.
\end{adjustwidth}

\section{Related Work}
 Unsupervised segmentation enabled by vision foundation models has made tremendous progress in recent years. In this section, we first introduce the relevant diffusion- and self-supervised learning-based~(SSL) foundation models and then discuss how these are leveraged for unsupervised segmentation.
\paragraph{Vision Foundation Models} that scale well with pre-training datasets, require few or no manually annotated samples, and generalize to new tasks or domains \citep{caron2021emerging, He2021MaskedAA} are often referred to as foundation models. Important components of this group of models are scalable architectures, mostly transformer-based such as the ViT \citep{dosovitskiy2021vit}, and effective pre-training tasks. Notable ones include contrastive learning \citep{caron2020unsupervised, chen20j, chen2021mocov3}, masked image modeling \citep{He2021MaskedAA,beit, eva02}, and self-distillation \citep{caron2021emerging, zhou2021ibot,oquab2023dinov2, simeoni2025dinov3}. By scaling challenging pre-text tasks, these models learn highly transferable features, especially for image perception \citep{oquab2023dinov2, kerssies2025eomt, cavagnero2026pmt}. Alongside SSL, generative vision models \citep{Rombach_2022_CVPR, esser2024scaling, Powell_ICLR2024} have emerged as a strong source of general features. \cite{tang2023emergent, meng2024not, stracke2025cleandift} found that single-timestep feature extraction is sufficient to get semantically coherent features, outperforming strong SSL baselines. The analysis by \cite{Zhang_2024_CVPR} further showed that generative models seem to learn strong representations for semantics in the value-value (v-v) interaction and instance discrimination in the query-key (q-k) interaction. To benefit from generalization and spatial grouping to generate dense instance masks, we also leverage the self-attention representation in our method \mymethod. Moreover, we mainly use SD2, a UNet-based architecture, and do not extract from diffusion transformer-based models. These currently have the drawback that the VAE latents are downsampled more strongly than in UNets, which limits the ability to discover small instances. In addition, they are computationally more demanding due to higher parameter count. Nevertheless, our method is similarly applicable to diffusion transformer affinities.

\paragraph{Unsupervised Instance Segmentation} approaches like TokenCut \citep{wang2022tokencut}, LOST \citep{LOST}, FOUND \citep{simeoni2023found}, and MOST \citep{Rambhatla2023MOSTMO} proposed methods to exploit the feature affinity of SSL pre-trainings as a graph structure and derived binary object masks based on threshold on the graph. To overcome the costly pseudo-label generation, \cutler proposed a simple training framework. At first, they labeled the ImageNet dataset with a modified NCut extended to multi-object segmentation and then trained an instance segmentation network on the labels for multiple rounds. CuVLer \citep{arica2024cuvler} improved this procedure by generating an ensemble of multiple pre-trained DINO models and creating majority vote labels. Several more advanced techniques were proposed, such as ProMerge \citep{li2024promerge}, which filtered feature similarity maps from prompts, CutS3D \citep{Sick_2025_ICCV}, which complemented the graph cut with depth, or UnMORE \citep{yang2025unmore}, which added a boundary reasoning stage on top of the CuVLer labels. One limitation of these approaches is that they operate at fixed granularities. Therefore, HASSOD \citep{cao2023hassod} proposed agglomeratively merging features at different thresholds, enabling hierarchies from sub-part to object. This allows a more general label generation. SOHES  \citep{cao2024sohes} extended HASSOD with different thresholds and an adapted training strategy, providing an unsupervised alternative to the widely used supervised SAM \citep{kirillov2023segany}. However, a significant performance gap remained that could be reduced by UnSAM. UnSAM \citep{wang2024segment} splits the task into detecting coarse objects by \cutler and clustering the finer parts of these objects with the algorithm introduced in HASSOD. Therefore, the approach can better handle high-resolution inputs and the tradeoff between parts/subparts and coarse instances. UnSAM thereby poses the state of the art in unsupervised instance segmentation.

\paragraph{Training-Free Segmentation with Generative Models}
Recently, it was shown that diffusion-based generative models 
can be utilized as generalist perception models.
In fact, they can be easily adapted for prompt-guided universal image and video understanding using paired image/video-text datasets through supervised fine-tuning \citep{gabeur2026image, wang2026diffusion, wang2026genception}. Specifically, a large body of work demonstrated that diffusion model representations can be leveraged for segmentation without retraining or supervision. Existing approaches broadly fall into two categories: methods that exploit image-text cross-attention to enable open-vocabulary zero-shot segmentation \citep{kim2025seg4diff, helbling2025conceptattention, sun2026iseg, Meng_2026_CVPR}, and 
methods that rely solely on internal model features for class-agnostic segmentation
\citep{tian2024diffuse, couairon2024diffcut, M2N2, Ivanova2024UnsupervisedSB, zhang2026falcon}. 
In this work, we focus on the latter setting and briefly review several representative approaches:
 DiffSeg \citep{tian2024diffuse}, for instance, employed simple agglomerative merging on the extracted self-attention and found that attention maps can be grouped into semantic segmentation clusters. Following up on that, DiffCut \citep{couairon2024diffcut} and Falcon \citep{zhang2026falcon} employ variations of the NCut and focus purely on feature-based affinities. However, all of these works mostly only profit from the semantic coherence of the model. TRACE \citep{jo2026trace} showed that diffusion models not only possess strong semantics but also meaningful edge information by extracting edge maps from self-attention that can post-process instance segmentation masks. M2N2 \citep{M2N2} extracts instance masks directly, showing that diffusion models support segmentation based on user clicks without training.  
Our work, \mymethod, however, differs from these approaches by leveraging
 generative models for \textit{automatic} mask generation to generate multi-granularity instance masks with dense coverage. Moreover, we address the limitations of existing work, such as M2N2, that requires user inputs and does not extract multiple granularities.

\section{\mymethod: Diffusion-Based 
Instance Segmentation Without Supervision}
In the following section, we first discuss the preliminaries of entity segmentation and diffusion-based feature extraction in \Cref{sec:method_preliminaries}, and provide a brief overview of our \mymethod approach in \Cref{sec:overview}. 
We then elaborate on the three main steps of \mymethod in more detail:
First, we explain the extraction of multi layer self-attention matrices and aggregation into an affinity matrix in \Cref{sec:attn_extract}. Second, in \Cref{sec:pseudo_label_gen}, we present the pseudo-label generation, consisting of automatic mask generation via non-linear propagation of point prompts into soft object masks using the affinity matrix, followed by mask merging and post-processing to obtain instance pseudo-labels at image resolution.  Finally, in \Cref{sec:training}, we describe training a lightweight model on the generated pseudo-labels to predict instance masks at multiple granularities.
\begin{figure}[!t]
    \centering
    \includegraphics[width=\linewidth]{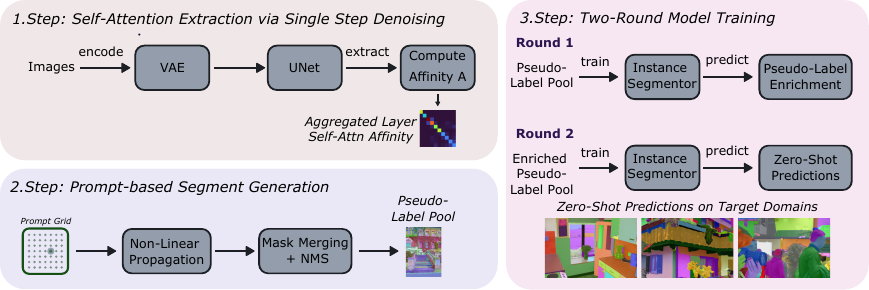}
    \caption{\textbf{Schematic depiction of our unsupervised \mymethod method.} \mymethod generates multi-granular instance masks, without supervision, from a pre-trained diffusion model in three steps. \textbf{Step~1:~self-attention extraction via single-step extraction} encodes an input image using the SD2 VAE and performs single-step denoising with the SD2 UNet to collect and aggregate multi-layer self-attention matrices into a single affinity matrix $\mathbf{A}$. \textbf{Step~2: prompt-based segment generation} initializes a grid of equidistant one-hot prompts and propagates them over the affinity into soft object maps using non-linear edge-stopping p-Laplacian propagation. We then produce multi-granular instance pseudo-labels through map merging, deduplication, and boundary refinement. \textbf{Step~3: two-round model training} trains a lightweight Mask2Former model on the generated pseudo-labels for one round, followed by self-training via crop-based prediction refinement to correct errors and discover fine-grained instances in the second round.}
    \label{fig:denoise2seg_overview}
\end{figure}
\subsection{Preliminaries}
\label{sec:method_preliminaries}
 \textbf{Segment Anything } introduced a model and dataset based on the entity segmentation definition \citep{qi2022open}. Its main novelty is a flexible prompt interface that takes points or boxes as input to segment coherent instances. Consistent with this definition, the training dataset SA-1B contains all kinds of objects and object parts, allowing the model to generalize across different label definitions and granularities. SA-1B is built with a complex data engine in which the SAM model is repeatedly retrained across three stages: first, initial manual labeling, second semi-automatic labeling with human assistance to fill blind spots, and third automatic mask generation from a grid of points, followed by a deduplication and filtering stage to produce high-quality annotations. With our method \mymethod, we mirror the latter but in an purely unsupervised setting. We place an evenly spaced grid of point prompts on the input image, generate candidate masks from the diffusion model's self-attention, then aggregate, filter, and post-process them into a final set of pseudo-labels. Inspired by SAM, we then train a segmentation model in a two-round procedure on the collected pseudo-labels to fill gaps and improve performance.

\paragraph{Generative Diffusion Models} are capable of learning complex data distributions through a two-directional process. Consider a data sample $\mathbf{x}_0 \sim p_{\mathrm{data}}$. The forward process is a Markov chain, gradually injecting Gaussian noise at each step,
$q(\mathbf{x}_t \mid \mathbf{x}_{t-1}) = \mathcal{N}\left(\sqrt{1-\beta_t}\,\mathbf{x}_{t-1},\, \beta_t \mathbf{I}\right)$,
with $t = 1, \dots, T$ and a chosen variance schedule ${\beta_t}$ resulting in $x_T$ being approximately Gaussian noise. A denoiser neural network $\boldsymbol{\epsilon}_\theta(\mathbf{x}_t, t)$ then learns to generate new samples by inverting the process and predicting the reverse transitions, $p_\theta(\mathbf{x}_{t-1} \mid \mathbf{x}_t)$. The denoiser is shared across all timesteps. To scale this process to high-resolution data, variational autoencoders (VAEs) were introduced for efficiency. Such a VAE maps the input image $I$ into a lower-dimensional latent representation $\mathbf{z} \in \mathbb{R}^{H_f \times W_f \times C}$ with an encoder, where $H_f \times W_f$ specifies the latent spatial resolution and $C$ the number of channels and reconstructs the input image again from $\mathbf{z}$ using a decoder. In current high-resolution image generation models, denoising is conducted only on the latent, usually using either a UNet or a diffusion transformer. Text-to-image models \citep{Rombach_2022_CVPR} additionally condition denoising on a text prompt $y$, allowing latent features to interact with text embeddings through cross-attention. As we target class-agnostic segmentation, we omit the text prompt and only pass null text embeddings. Consequently, we restrict ourselves to image-to-image attention and ignore text-to-image cross-attention in our experiments.  We primarily adopt the UNet-based SD2 architecture, whose latent is downsampled less aggressively compared to diffusion transformers. This enables us to better handle small instances and preserve fine details. Since the denoising UNet is trained to denoise at different levels $t = 1, \dots, T$, it can be repurposed as a feature extractor for images running just a single step \citep{luo2023diffusion, meng2024not, Zhang_2024_CVPR, stracke2025cleandift}. This involves encoding the input image into the latent representation $\mathbf{z}$ without applying the forward noise. During denoising, we can cache and leverage internal representations for downstream tasks. In our case, we are interested in high-resolution self-attention. Therefore, we cache multi-layer self-attention outputs from the last UNet block at the highest resolution. Depending on the timestep, different features are produced. Thus, the feature extraction can be optimized for specific target tasks.

\subsection{Overview of \mymethod}\label{sec:overview}
Our method \mymethod leverages concepts from both SAM and generative diffusion models to densely label images in a unsupervised fashion. It is inspired by SAM's automatic mask generation and self-training for high-quality label generation, which we mimic in a purely unsupervised setting. Specifically, \mymethod consists of three steps (see \Cref{fig:denoise2seg_overview} for a schematic visualization): 
First, in step~1, the self-attention single-step extraction, we conduct single-step denoising to extract and aggregate self-attention matrices from a pre-trained diffusion model into a single affinity matrix. %
Next, in step~2, pseudo-label generation, we utilize a grid of equidistant one-hot point prompts to generate soft object maps using the affinity matrix. 
Subsequently, we isolate multi-granularity instances, also referred to as pseudo-labels, from these soft object maps using mask merging and non-maximum suppression. 
Consistent with prior work, we apply a mask refinement step to improve the quality of mask boundaries. 
Finally, in step~3, training on pseudo-labels, we train a segmentation network on the generated pseudo-labels. This training includes two rounds: first, a training on the generated pseudo-labels, followed by a second round on prediction-enhanced pseudo-labels to fill missing segments and correct model errors. In the following, we describe the three steps of \mymethod in greater detail.

\subsection{Step 1: Self-Attention Extraction via Single-Step Denoising}
\label{sec:attn_extract}
For our pseudo-label generation pipeline, we start by extracting self-attention matrices from the StableDiffusion2 (SD2) UNet \citep{Rombach_2022_CVPR}. 
To do so, we encode an input image using the SD2 VAE encoder to obtain its latent representation $\mathbf{z}$, run a single step of denoising and cache the obtained self-attention matrices. %
Specifically, for each layer $l \in\{1, ..., n_L\}$ and attention head $h \in\{1, ..., n_H\}$, we store one $N \times N$ matrix $\textbf{A}^{(l,h)}$ that represents the affinity between the N patch tokens.
Following \citet{M2N2}, we use only the self-attention tensors from the first two layers at the highest resolution of the UNet decoder. We first aggregate each layer over its heads and then combine layers via weights $w_l$. The weights $w_l$ sum to 1 and are treated as hyperparameters (see \Cref{sec:app_nonlocal}). The scene representation of the diffusion model is condensed into the aggregated affinity matrix used for the subsequent steps of our label generation. The final aggregated self-attention matrix then takes the form
\begin{equation}
    \textbf{A} = \frac{1}{n_h}\sum_{l=1}^{n_L} w_{l} \sum_{h=1}^{n_H} \textbf{A}^{(l,h)}\ .
\end{equation}
\subsection{Step 2: Prompt-Based Segment
Generation}
\label{sec:pseudo_label_gen}
\begin{figure}
    \centering
    \includegraphics[width=\linewidth]{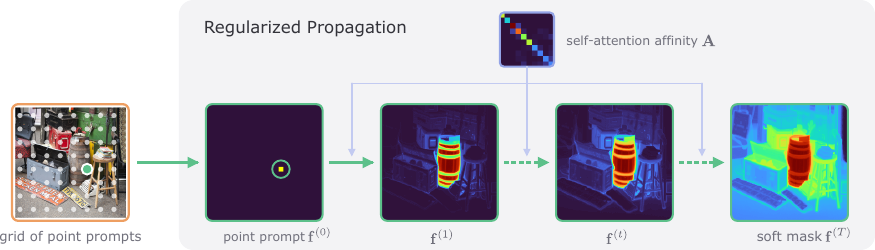}
    \caption{\textbf{Generation of soft object maps by prompt propagation.} An equidistant grid of one-hot point prompts is placed across the latent token grid $(H_f \times W_f)$ as initial seeds (grey dots). Each prompt is propagated over the self-attention affinity graph $\textbf{A}$ via edge-preserving non-linear p-Laplacian propagation (\Cref{optimeq}) until it reaches $\|\mathbf{f}^{t+1} - \mathbf{f}^{t}\|_2^2  \leq \tau_{\text{prop}}$, yielding continuous soft object maps. In this example, we show the propagation of a single prompt (green dot).}
    \label{fig:placeholder2}
\end{figure}
  Self-attention in generative diffusion models concentrates on semantically similar regions and encodes object boundaries \citep{tian2024diffuse, M2N2, erel2025attention, jo2026trace, zhang2026anchordiff}. Which are two useful properties for isolating instances. However, simply thresholding the affinity maps, as done in \citet{li2024promerge}, is insufficient because the self-attention matrices are sparse. Instead we treat the aggregated affinity matrix $\mathbf{A}$ as a graph for prompt propagation. This yields dense and coherent object maps as shown in the following steps.

\paragraph{Prompt Propagation} %
We build on the non-local graph regularization of \citet{Elmoataz08}, which adaptively smooths an input signal while preserving edges. In our case, these discontinuities represent the attention differences on object edges. The smoothing process is controlled by the affinity extracted in step 1: it enforces neighboring tokens to take similar values for high-affinity neighbors while throttling the smoothing effect for low-affinity neighbors. This amplifies propagation within objects and limits it at edges. To segment the whole 
image, we generate a grid of equidistant prompts with fixed spacing on the latent representation. %
Each prompt covers a different image region and is represented by a one-hot vector $\mathbf{f}^0 \in \mathbb{R}^N_{+}$. Then we smooth these prompts into soft object maps $\mathbf{f} \in \mathbb{R}^N_{+}$ by minimizing

\begin{equation}
E_{A}^p(\mathbf{f}, \mathbf{f}^0, \lambda) \ =\  \frac{1}{p} \sum_{i=1}^{N}
\left( \sum_{j=1}^{N} A_{ij}(f_j - f_i)^2 \right)^{p/2}+\ \ \frac{\lambda}{2} ||\mathbf{f} - \mathbf{f}^0||_2^2
\ \ 
\label{optimeq}
\end{equation}
with respect to $\mathbf{f}$. This is done for all prompts from the grid independently and 
in parallel. The first term smooths a prompt $\textbf{f}$ using the affinity matrix, and the second term regularizes the smoothing by anchoring it to the initial prompt. The exponent $p$ controls how strongly discontinuities are preserved, where $p=2$ is equivalent to a linear propagation,  and $\lambda$ governs the anchoring to the initial prompt. Following \citet{Elmoataz08}, we solve this optimization problem efficiently with Gauss-Jacobi iterations (see \Cref{sec:app_nonlocal} for details) and stop when the squared norm of the iteration update $ \|\mathbf{f}^{t+1} - \mathbf{f}^{t}\|_2^2$ falls below $\tau_{\rm prop}$, producing our final set of soft object maps.

\paragraph{Map Merging}
These soft object maps are often redundant or capture the same object at different granularities. Thus, we consolidate them into a set of complementary multi-granularity 
masks with the following steps: First, we normalize the maps into valid probability distributions $\textbf{p}_{k} = \textbf{f}_{k}/{\sum_i f_{k,i}}$. %
We then measure the dissimilarity of a prompt $k$ with another prompt $k'$ using the symmetric Kullback-Leibler (KL) divergence
\begin{equation}
    d_\mathrm{KL} = \frac{1}{2} \biggl( D_\mathrm{KL}\Bigl(\mathbf{p}_{(k)} \!\parallel\! \mathbf{p}_{(k')}\Bigr) + D_\mathrm{KL}\Bigl(\mathbf{p}_{(k')} \!\parallel\! \mathbf{p}_{(k)}\Bigr) \biggr)\ .
    \label{eq:sym_kl}
\end{equation}
Based on these distances, we cluster the prompts with agglomerative average linkage clustering on $\textbf{D}$. The distance matrix $\textbf{D}$ here contains the pairwise-distances for all prompts and is of shape $ K \times K$. Cutting the resulting clustering dendrogram at different threshold values $h$ groups the prompts into clusters $\mathcal{C}^h$. Small thresholds $h$ result in a larger number of clusters and are thus suited to isolate small objects, whereas larger h values yield fewer clusters and are thus favorable for isolating whole or coarse objects. To infer high-resolution instance masks from the merged clusters, we average all maps $\textbf{p}_c$ belonging to a cluster $C$ in the set of clusters $\{\mathcal{C}^h\}_h$ at a threshold $h$ with  $\bar{\textbf{p}}_{C} = \sum_{c \in \mathcal{C}} {\textbf{p}}_{c}/C$. Then we upsample the merged maps and perform argmax on them to obtain a cluster segmentation map $\textbf{M}^h$ for each level $h$. Finally, we isolate instances out of the high-resolution segmentation by connected components \citep{Bolelli2020Spaghetti}. Consequently, we obtain connected regions forming our instance masks. To %
reduce noise,
we filter out components with an area below $a_{min}$ (for detailed pseudo-code, see \Cref{sec:app_nonlocal}). We repeat this procedure for all clustering thresholds to add multi-granularity segmentation proposals to the candidate pool $\mathcal{P}$.

\textbf{Mask Post-Processing}
Since adjacent clustering thresholds $h$ produce similar segmentation masks, the candidate pool $\mathcal{P}$ 
We remove redundant masks via a simple area-descending non-maximum suppression.
Candidate masks are sorted in descending order of pixel area and added to the final annotation set $\mathcal{M}$.
A candidate mask $m$ is rejected if any already-accepted mask $m' \in \mathcal{M}$ satisfies $\text{IoU}(m, m') > \tau_{\text{IoU}}$.
This procedure creates a set $\mathcal{M}$ of at most $N_{\max}$ complementary masks per image, where $N_{\max}$ represents the maximum amount of pseudo masks per image and is set to 1000. Similar to baselines such as UnSAM \citep{wang2024segment} and SOHES \citep{cao2024sohes}, we employ CascadePSP \citep{cheng2020cascadepsp} as an optional refinement step on the generated masks.

\subsection{Step 3: Two-Round Model Training on Generated Segments
}
\label{sec:training}
As the previously outlined pseudo-label generation requires inference of the pre-trained diffusion model, generating masks for new images can be costly. Moreover, the pseudo-labels are not perfect, as the pre-trained diffusion model is not an expert for instance segmentation. Therefore, we train a lightweight segmentation model on the pseudo-masks from the previous pseudo-label generation step using different image budgets that range from 1\% to 4\% of the full training dataset. We train for two consecutive rounds, similar to the multi-round training of SAM, such that the model can improve label quality by self-training on its own predictions. The training follows the prior work UnSAM and is based on the Mask2Former architecture \cite{cheng2022mask2former}, which is optimized in a class-agnostic manner with the loss

\begin{equation}
    \mathcal{L} = \lambda_{\textrm{dice}} \,\mathcal{L}_{\textrm{dice}} + \lambda_{\textrm{bce}}\,\mathcal{L}_{\textrm{bce}} + \lambda_{\textrm{cls}}\,\mathcal{L}_{\textrm{cls}}\ .
\end{equation}

The loss terms $\mathcal{L}_{\textrm{dice}}$ and $\mathcal{L}_{\textrm{bce}}$ optimize for mask quality whereas $\mathcal{L}_{\textrm{cls}}$ guides the model towards learning an objectness probability. The loss contributions are weighted via the respective hyperparameters. 
In the following, we present two different configurations of this two-round training procedure. The first one is based purely on the diffusion-based labels, whereas the second one follows the prior state-of-the-art UnSAM and augments 
the set of generated masks
with masks from a pre-trained \cutler segmentation model. 

\paragraph{Round~1: Training on Generated Segments} 
For the first round, both configurations take the same pool of previously generated segments as labels.
As the latent resolution of SD2 UNet is limited, we lack some small-scale segments and thus need to increase the frequency and diversity of these segments during training. This is done by applying copy-and-paste augmentation \citep{Ghiasi2020SimpleCI}. In our detector-augmented setting, we combine small-object pasting with predictions from \cutler, with a confidence above $\tau_{\rm CutLER}$, set to 0.1 throughout experiments. This ensures 
a sufficient number of labels across coarse instances, parts, and sub-parts. Moreover, we make sure that not a single segment source dominates by randomly deciding the source of supervision at each iteration: i) both \cutler and \mymethod masks with a probability of 50 \% each, ii) only \cutler masks with 20 \%, and iii) only \mymethod masks with 30\%. 

\paragraph{Round~2: Training with Predict-and-Conquer Enrichment}
In the second round, we use the model trained in Round 1 to enrich the pseudo-label pool. Round 1 labels are spatially dense, but they miss fine details because of the limited latent resolution of the SD2 UNet. Inspired by UnSAM’s divide-and-conquer strategy, we therefore combine a full-image inference pass with a second pass on crops of previously predicted parent segments. We first run the Round 1 model on the whole image and keep every prediction with a score of at least $\theta$. From this set, we discard near-full-image predictions ($> a_{\rm high}$) and very small predictions ($< a_{\rm low}$), which are unlikely to contain smaller parts. A child is kept only if it is contained in its parent, as $
\mathrm{area}(\mathrm{child}\cap\mathrm{parent})/\mathrm{area}(\mathrm{child})\ge\tau_{\mathrm{cont}}$, so that we do not simply re-detect the parent. The surviving global and crop predictions $\hat{m}$ are then merged into the initial Round 1 pool: any original mask $m$ with $\text{IoU}(m, \hat{m}) > \tau_{\rm merge}$ is replaced, and the remaining predictions are appended.
This merge is the same for both training configurations. They differ only in whether CutLER masks are added afterward. In the detector-free setting, Round~2 trains on the merged pool above (initial \mymethod labels plus predict-and-conquer predictions). In the detector-augmented setting, we additionally append CutLER masks to that pool and enable hierarchical copy-paste of CutLER instances during training. Crop refinement itself never depends \mbox{on CutLER}.

\section{Experiments}
In this section, we evaluate our entity-segmentation method \mymethod on a variety of target datasets. First, we outline the experimental setup for pseudo-label generation and training in \Cref{subsec:exp_setup}. Next, we compare our pseudo-label generation against SSL-based and diffusion-based label-generation baselines in \Cref{sec:pseudo-label-quality}. Subsequently, we train a segmentation network on our pseudo labels and report its zero-shot performance in \Cref{sec:pseudo-label-training}. Next, we analyze using our trained model as initialization for efficient semi-supervised fine-tuning on labeled data in \Cref{sec:semisupervised}. Finally, we analyze the design choices of \mymethod in a series of ablations in \Cref{sec:ablations}.

\subsection{Experimental Setup}\label{subsec:exp_setup}
In this subsection, we first describe the datasets used to train and evaluate \mymethod. Then, we outline the SSL-based and diffusion-based benchmark methods against which we compare \mymethod. Finally, we present the 
details of pseudo-label generation and training and the metrics used to evaluate segment quality.
\paragraph{Datasets} We use different dataset setups for evaluating pseudo-label quality and model training with zero-shot inference. We perform pseudo-label generation for training on randomly sampled subsets of the SA-1B dataset, comprising 100k, 200k, and 400k images, which correspond to approximately 1\%, 2\%, and 4\% of the entire SA-1B training dataset. We also sample 1000 additional non-overlapping images for validation, which we use throughout all experiments. To investigate our method's ability to annotate images densely, we moreover select several additional open-world validation datasets. These include ADE20K \citep{zhou2017scene}, EntitySeg \citep{Qi_2023_EntitySeg}, and UVO \citep{Wang_2021_ICCV} for segmenting entities. Furthermore, we analyze the part segmentation performance on PACO \citep{Ramanathan_2023_CVPR}. To evaluate our trained model, we also add COCO Images \citep{lin2014microsoft} to cover “things”-only performance. We evaluate both the COCO and the more diverse LVIS validation annotations. Compared to prior work, we intentionally do not evaluate on PartImageNet as this dataset contains images from ImageNet, the pre-training dataset for both DINO and CutLER. Benchmarking on this dataset is thus not fair for comparing SD2- and DINO-based approaches.
\paragraph{Baseline Methods} We evaluate our pseudo-labeling mechanism against common baselines for unsupervised segmentation. For SSL-based unsupervised segmentation, DINO pre-training poses a widely used standard. Additionally, we consider entity-segmentation baselines, including SOHES \citep{cao2024sohes}, which performs global clustering and local re-clustering, and UnSAM, which relies on divide-and-conquer that first discovers coarse regions and then clusters them bottom-up. 
 For diffusion-based pre-training, there is no prior work on open-world entity segmentation. That is why we repurpose several established baselines in unsupervised semantic segmentation. DiffSeg \citep{tian2024diffuse} uses a similar approach and merges  attention maps into segmentation masks. We adapt the work by applying connected components at the end to isolate instances. Moreover, we add the recent method M2N2 \citep{M2N2}, which repurposes generative diffusion models for interactive segmentation. It propagates user-defined point prompts into segmentation masks. To make a fair comparison, we apply the same grid of point prompts and extract the resulting masks at image resolution. 
 For our training experiments, we compare to UnSAM, the state-of-the-art for unsupervised entity segmentation, and, additionally, to established unsupervised instance segmentation methods that incorporate training, such as SOHES and \cutler \citep{wang2023cut}.
\paragraph{Pseudo-Label Settings}
For our pseudo-label creation, we mainly use the SD2 model. We perform one step of denoising at timestep 150 following \citet{M2N2} and resize the input images to $1120 \times 1120$, as we found that SD2 generalizes to resolutions slightly above its training resolution of $1024 \times 1024$. Furthermore, we utilize the aggregation weights of $w_1 = 0.85$ and $w_2 = 0.15$ for the last two layers at the highest resolution of SD2's U-Net. We only use the layers at the highest resolution to retain fine-grained image details. On our final self-attention affinity we apply temperature scaling with $\tau_{att}=0.55$ to soften propagation. For our prompting, we set the distance between prompts to six pixels horizontally and vertically. In addition, we set the propagation gradient exponent to $p=1.6$, and the anchoring penalty $\lambda = 1 \times 10^{-5}$, as we found these values to balance small-, medium-, and large-object recall well and that prompt anchoring is only needed as slight regularization. We iterate until we reach the stopping criterion $\tau_{\rm prop} = 1 \times 10^{-4}$ to preserve segmentation mask granularity. For merging, we compute six thresholds in the range [0.186, 2.99] that are log-spaced to emphasize merges at lower thresholds. The area-based deduplication is done with the geometric merging parameter $\tau_{\rm\,IoU} = 0.9$, which worked best to preserve recall. We remove all masks containing fewer than 100 pixels to remove noise. We performed ablations on SA-1B to validate the behavior of our pseudo-labeling parameters (see \Cref{sec:app_nonlocal}). For post-processing, we 
largely adopt the CascadePSP parameters of UnSAM and SOHES; for details, see \Cref{sec:cascadepsp_hyperarameters}. For training we mainly follow the hyperparameters of UnSAM, as detailed in \Cref{sec:appendix_train_hyperarameters}.

\paragraph{Pseudo-Label and Zero-Shot Metrics}
To evaluate both the quality of the diffusion-based pseudo-labels and the zero-shot performance of the derived entity-segmentation models, we mainly consider mean average recall ($\text{AR}_{1000}$/mAR), which is established in open-world entity segmentation \citep{wang2023cut, cao2024sohes, wang2024segment}. It averages recall over IoU matching thresholds from 0.5 to 0.95. We always consider up to 1000 predicted masks per image for metric calculation. In addition, we report average recall filtered by object size, denoted as $\mathrm{AR}_{S}$, $\mathrm{AR}_{M}$, and $\mathrm{AR}_{L}$, following the COCO \citep{lin2014microsoft} definition. Because label definitions and taxonomies vary across datasets, mAP is less meaningful and is only considered for selected experiments. 

\subsection{Evaluation of Generated Pseudo-Labels}
\label{sec:pseudo-label-quality}
To evaluate the quality and generalization of our diffusion-based pseudo-labels, we first analyze their performance on SA-1B in \Cref{tab:pseudo-quality-sa1b}, comparing them against the baselines described in \Cref{subsec:exp_setup} without mask-refinement post-processing. 
Labeling the high-resolution SA-1B images is particularly challenging, as most methods operate on latent representations with substantially lower resolution. Nevertheless, despite operating on such a low-resolution representation, our method achieves higher average recall than all baselines:
\mymethod's mAR is 4.7 p.p.\ higher than the strongest diffusion-based baseline DiffSeg \citep{tian2024diffuse} and 4.3 p.p.\ higher than the prior DINO-based state-of-the-art UnSAM \citep{wang2024segment}. 

Comparing the diffusion-based baselines, we find DiffSeg to significantly outperform M2N2 at all object scales, likely because its agglomerative clustering covers multiple object granularities. In contrast, M2N2 \citep{M2N2} only produces a single granularity mask per prompt. For the DINO-based baselines, prior state-of-the-art UnSAM has an advantage on small objects: its divide-and-conquer strategy uses predictions from the pre-trained CutLER instance segmenter and 
further splits the CutLER-based instances into smaller parts using a bottom-up approach.
However, its dependency on the ImageNet pre-trained \cutler detector can cause the method to miss some coarse objects and their parts entirely if the object is not detected or assigned low objectness. This lowers UnSAM’s recall on medium and large objects relative to our approach, which avoids this dependency for pseudo-label generation.
\begin{table}[htb]
    \tablestyle{4pt}{1.05}
    \begin{center}
    \begin{tabular}{lcrrrr}
        \toprule
        \textbf{Method} & \textbf{Backbone} & \textbf{AR}$_S$ & \textbf{AR}$_M$ & \textbf{AR}$_L$ & \textbf{AR}$_{1000}$ \\
        \midrule
        \addlinespace[4pt]
        \multicolumn{6}{l}{\textit{DINO-based labeling}} \\[2pt]
        UnSAM ~\citep{wang2024segment} & DINO ViT-B/8 & \textbf{5.7} & 13.9 & 26.0 & 16.4 \\
         SOHES~\citep{cao2024sohes} & DINO ViT-B/8 & 2.8 & 10.6 & 15.7 & 11.0 \\
        \midrule
        \addlinespace[4pt]
        \multicolumn{6}{l}
        {\textit{Diffusion-based labeling}} \\[2pt]
        DiffSeg~\citep{tian2024diffuse} & SD2 & 0.3 & 12.4 & 29.9 & 16.0 \\
        M2N2~\citep{M2N2} & SD2 & 0.0 &  2.8 & 26.5 & 9.7 \\
        \rowcolor{gray!10}
        \mymethod (\textbf{ours}) & SD2 & 2.6 & \bf 15.3 & \bf 38.9 & \bf 20.7 \\
        \bottomrule
    \end{tabular}
    \end{center}
    \caption{\textbf{Comparison of DINO- and diffusion-based pseudo-labeling methods on the SA-1B validation set.} We report class-agnostic mask average recall (AR) by object size, with 1000 proposals per image. Methods are grouped by backbone type. 
    The diffusion-based labels of \mymethod achieve significantly higher average recall compared to previous methods.}
    \label{tab:pseudo-quality-sa1b}
\end{table}

Next, we compare these segmentation methods on datasets from diverse domains, as shown in \Cref{tab:pseudo-quality-crossdomain}, keeping the previously used hyperparameter values. This cross-domain analysis covers a broad range of datasets, ranging from ADE20k that addresses Scene Parsing with a huge number of different classes, over EntitySeg that encompasses diverse entities from the open world, to UVO that provides densely annotated frames from unstructured web videos, and PACO which targets part segmentation.
Among the training-free methods, our \mymethod technique clearly benefits from the generality of diffusion-model representations and consistently achieves the highest recall, reaching, for example, an $\text{AR}_{1000}$ of 13.6 on PACO, a 2.9~p.p.\ margin over the strongest baseline \cutler, despite the latter being trained. This demonstrates \mymethod's ability to generate multi-granularity masks. In addition, we  outperform the training-free baselines on entity-focused datasets such as ADE20k and EntitySeg, by a large margin gaining 6.5~p.p.\ and 4.5~p.p.\ of mAR compared to the prior state-of-the-art technique UnSAM. Generally, our diffusion-based approach outperforms DINO-based approaches in this setup, confirming our observation that diffusion priors are better suited for segmenting stuff+things datasets than DINO features. We further show improvements across domains compared to other diffusion-based baselines. M2N2 
performs weakly
on datasets with objects at different granularities and is roughly on par with \mymethod on UVO only due to the dataset's whole-object focus. DiffSeg, in contrast, includes a mechanism to cover different granularities. However, 
without a propagation step, it appears to be much less stable across domains, with a gap of up to 11.0~p.p.\ to our method. Notably, our method outperforms the CutLER-trained detector, which was trained on 1.3M ImageNet images for segmentation, on SA-1B, EntitySeg, and PACO and stays close on UVO and ADE20k. 
\begin{table}[htb]
    \tablestyle{4pt}{1.05}
    \begin{center}
    \begin{tabular}{lcrrrrr}
        \toprule
        \multirow{2}{*}{\textbf{Method}} & \multirow{2}{*}{\textbf{Backbone}} & \multicolumn{5}{c}{\textbf{AR$_{1000}$ Across Domains}} \\
        \cmidrule(lr){3-7}
        & & SA-1B & ADE20K & EntitySeg & UVO & PACO \\
        \midrule
        \addlinespace[5pt]
         \multicolumn{6}{l}{\textit{Training-based labeling}} \\[2pt]
         CutLER ~\citep{wang2023cut} &  DINO R50 & 17.0  &  24.8 &  22.1 & 32.3 &  10.7 \\
          \midrule
        \addlinespace[5pt]
        \multicolumn{6}{l}{\textit{DINO-based labeling}} \\[2pt]
        \color{black} UnSAM ~\citep{wang2024segment} & DINO ViT-B/8 & 16.4 & 16.0 & 19.3 & 22.8 & 9.3 \\
        SOHES~\citep{cao2024sohes} & DINO ViT-B/8 & 11.0 & 11.6 & 11.6 & 19.3 & 5.2 \\
        \midrule
        \addlinespace[4pt]
        \multicolumn{6}{l}{\textit{Diffusion-based labeling}} \\[2pt]
        DiffSeg~\citep{tian2024diffuse} & SD2 & 16.0 & 14.1 & 17.5 & 18.9 & 9.8 \\
        M2N2~\citep{M2N2} & SD2 & 9.7 & 18.9 & 19.7 & 29.7 & 9.6\\
        \rowcolor{gray!10}
        \mymethod (\textbf{ours}) & SD2 & \textbf{20.7} & \textbf{22.5} & \textbf{23.8} & \textbf{29.9} & \textbf{13.6} \\
        \bottomrule
    \end{tabular}
    \end{center}
    \caption{\textbf{Cross-domain pseudo-label quality}, as measured by class-agnostic mask AR$_{1000}$ against the ground truth of the respective dataset. Apart from SA-1B, we probe generalization to scene parsing (ADE20K), entity segmentation (EntitySeg), and open-world video objects (UVO). The diffusion features produce more robust labels across domains, as indicated
    by consistently higher recall values. We keep for all methods the same hyperparameters as for SA-1B. We mainly focus on the performance of the training-free approaches and report \cutler because of its integration into UnSAM and as a widely used training-based reference.}
    \label{tab:pseudo-quality-crossdomain}
\end{table}
\subsection{Evaluation of Zero-Shot Instance Segmentation}
\label{sec:pseudo-label-training}
\begin{table*}[htb]
    \tablestyle{4pt}{1.05}
    \begin{center}
    \begin{tabular*}{0.97\textwidth}{
        @{\extracolsep{\fill}}
        lccrr>{\columncolor{gray!15}}r
        rrr>{\columncolor{gray!15}}r r
    }
        \toprule
        \multirow{2}{*}{\textbf{Method}} & \multirow{2}{*}{\textbf{\makecell{Detector\\Supervision}}} & \multirow{2}{*}{\textbf{\#\thinspace Images}} &
        \multicolumn{3}{c}{\textbf{Things}} & \multicolumn{4}{c}{\textbf{Stuff + Things}} & \textbf{Parts} \\
        \cmidrule(lr){4-6} \cmidrule(lr){7-10} \cmidrule(lr){11-11}
        &&& COCO & LVIS & \cellcolor{white} Avg. & ADE & Entity & SA-1B & \cellcolor{white} Avg. & PACO \\
        \midrule[1.5pt]
        \addlinespace[5pt]
        \multicolumn{11}{l}{\textit{Supervised}} \\[3pt]
        SAM & \xmark & 11M & 49.6 & 46.1 & 47.9 & 45.8 & 45.9 & 60.8 & 50.8 & 18.1 \\
        \midrule
        \midrule
        \addlinespace[5pt]
        \multicolumn{11}{l}{\textit{Unsupervised - label generation without a pre-trained detector}} \\[3pt]
        CutLER & \xmark & 1.3M & 28.1 & 20.2 & 24.2 & 26.3 & 23.1 & 17.0 & 22.1 & 8.9 \\
        SOHES & \xmark & 0.2M & 30.5 & 29.1 & 29.8 & 31.1 & 33.5 & 33.3 & 32.6 & 17.1 \\
        \mymethod (\textbf{ours}) & \xmark & 0.1M & \textbf{37.4} & \textbf{36.9} & \textbf{37.2} & \textbf{37.6} & \textbf{39.7} & \textbf{43.5} & \textbf{40.3} & \textbf{27.1} \\
        \midrule
        \addlinespace[5pt]
        \multicolumn{11}{l}{\textit{Unsupervised - label generation with a pre-trained detector (CutLER proposals)}} \\[3pt]
        \ours & \cmark & 0.1M & 40.5 & 37.7 & 39.1 & 35.7 & 39.6 & 41.9 & 39.1 & 27.5 \\
        \mymethod (\textbf{ours}) & \cmark & 0.1M & \textbf{41.0} & \textbf{38.7} & \textbf{39.9} & \textbf{39.8} & \textbf{42.2} & \textbf{45.5} & \textbf{42.5} & \textbf{28.9} \\
        \midrule
        \addlinespace[5pt]
        \ours & \cmark & 0.2M & \textbf{41.2} & \textbf{39.7} & \textbf{40.5} & 36.8 & 40.3 & 43.6 & 40.2 & 29.1 \\
        \mymethod (\textbf{ours}) & \cmark & 0.2M & 40.7 & 39.0 & 39.9 & \textbf{39.6} & \textbf{42.2} & \textbf{46.6} & \textbf{42.8} & 29.1 \\
        \midrule
        \addlinespace[5pt]
        \ours & \cmark & 0.4M & 42.0 & \textbf{40.5} & \textbf{41.3} & 39.1 & 41.1 & 44.8 & 41.7 & 29.7 \\
        \mymethod (\textbf{ours}) & \cmark & 0.4M & \textbf{42.2} & 40.0 & 41.1 & \textbf{40.5} & \textbf{43.2} & \textbf{47.8} & \textbf{43.8} & \textbf{29.8} \\

    \end{tabular*}
    \end{center}
    \caption{\textbf{Comparison of instance segmentation methods with and without auxiliary detector supervision.} While the previous state-of-the-art method UnSAM relies on an ImageNet pre-trained detector, our method can also be trained without auxiliary supervision purely on 
    diffusion-model-based labels. Combining our method with CutLER supervision positions it as complementary to the object-biased labels of UnSAM. In fact, dense supervision leads to significant improvements on stuff+things datasets while performing on par on things-only datasets.}
    \label{tab:results-ar-by-budget}
\end{table*}
We now evaluate the final \mymethod instance segmentation models trained with pseudo-label supervision and compare them with unsupervised segmentation baselines trained either with or without detector supervision (\Cref{tab:results-ar-by-budget}). All models are optimized on SA-1B and inferred zero-shot on previously unseen datasets.

When comparing \mymethod to SOHES, the strongest detector-free baseline, our method provides large gains on every benchmark: +7.4 p.p.\ AR on things datasets (37.2 vs.\ 29.8), +7.7 p.p.\ on stuff+things (40.3 vs.\ 32.6), and +10.0 p.p.\ on parts benchmarks (27.1 vs.\ 17.1). These results show that our diffusion-based approach is significantly more data-efficient, requiring only half the training images and a much smaller image encoder (ResNet50 for \mymethod vs.\ ViT-B for SOHES).

For training with detector supervision, we compare to the prior state-of-the-art technique UnSAM, which bases its label generation on ranked proposals from a pre-trained \cutler object detector. If we augment our segment label set with the same proposals, we consistently surpass UnSAM on stuff+things datasets.

Across all training image budgets ranging from 1\% over 2\% to 4\% of the SA-1B dataset, we improve the mAR by +3.4 p.p., 2.6 p.p., and 2.1 p.p., respectively. On the things and parts datasets, we perform on par with UnSAM, which performs very well on things-focused datasets due to its deep integration and learned proposal ranking of the CutLER detector. Overall, we show that our diffusion-based method \mymethod offers strong stuff+things coverage, \begin{wrapfigure}{r}{0.42\linewidth}
  \centering
  \includegraphics[width=\linewidth]{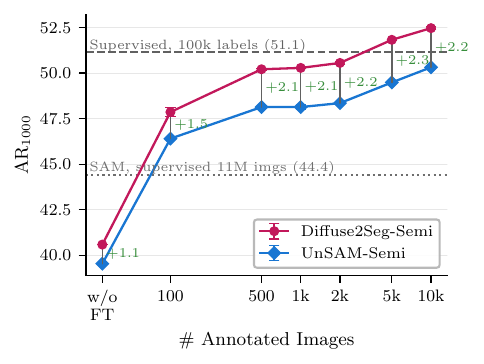}
  \caption{\textbf{Semi-supervised extensions of UnSAM and \mymethod.}
  Mean average recall vs.\ the number of labeled fine-tuning images. 5k images suffice for \mymethod to match the 100k supervised UnSAM reference.}
  \label{fig:label-efficiency-ar}
 \vspace{-1.2cm} 
\end{wrapfigure} surpassing the state-of-the-art DINO-based UnSAM. 
\paragraph{Comparison to supervised SAM}
As expected, a certain performance gap to supervised SAM remains, as SAM was trained on many more images (11M) and for many more rounds. Nevertheless, our method 
approaches SAM performance on
generalizable stuff+things segmentation, being just 2.7 p.p.\ worse in mAR on EntitySeg than SAM.
\subsection{Semi-Supervised \mymethodplus 
}\label{sec:semisupervised}
A central promise of unsupervised segmentation is that it reduces the need for costly labeling.
So far, we have shown that \mymethod produces strong segmentation models in a purely unsupervised setting. While a gap remains relative to supervised SAM, a natural next question is whether we can benefit from our method when manual labeling of smaller amounts of images is feasible. Therefore, we study label-efficient semi-supervised learning by fine-tuning our best unsupervised model and UnSAM's best unsupervised model with a simple recipe at a 4\% SA-1B budget as indicated in \Cref{fig:label-efficiency-ar}. We freeze the encoder to preserve generalization and unfreeze the decoder to align its mask predictions with the ground truth. As a strong supervised reference, we additionally train the same UnSAM recipe from scratch on 100k labeled samples. This fully supervised model reaches 51.1 mAR, exceeding the SAM reference at 44.4 mAR, because its design is tailored towards automatic mask generation. Fine-tuning our model is remarkably data-efficient: With just 100 samples available, our model already recovers 93.5 \% of the mAR of the fully supervised 100k reference, compared to 90.7 \% for UnSAM under the same protocol.
In this setup, the advantage of our semi-supervised Diffuse2Seg further grows with increasing image budget.  At just 5,000 examples, \mymethodplus matches the supervised reference performance, and at 10,000 we outperform fully supervised learning by 1.4 p.p. in average recall across all domains. This shows that our \mymethodplus model is not only effective in the unsupervised regime but can moreover significantly reduce the amount of labeled data required to outperform fully supervised benchmarks.

\subsection{Ablation Study}\label{sec:ablations}

\begin{figure}[t]
  \centering
    \begin{subfigure}[t]{0.48\linewidth}
        \centering
        \includegraphics[width=\linewidth]{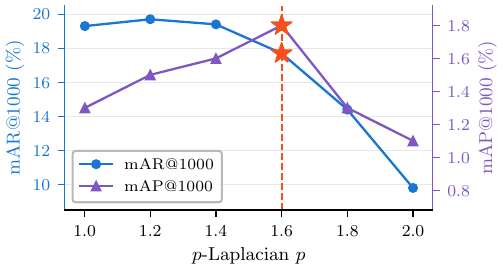}
        \caption{Influence of propagation exponent $p$.}
        \label{fig:parameter-behavior-p}
    \end{subfigure}\hfill
    \begin{subfigure}[t]{0.48\linewidth}
        \centering
        \includegraphics[width=\linewidth]{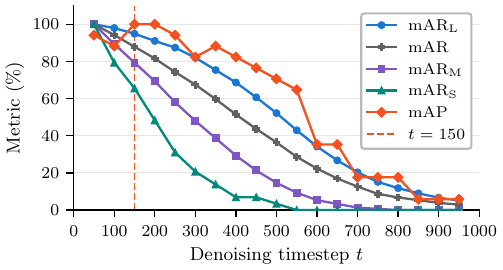}
        \caption{Behavior of the timestep t for extraction.}
        \label{fig:parameter-behavior-timestep}
    \end{subfigure}
    \caption{\textbf{Impact of propagation exponent and denoising timestep on \mymethod performance}. Panel~\subref{fig:parameter-behavior-p}: mAR@1000 peaks at $p=1.2$, while mAP@1000 peaks at the selected $p=1.6$. Panel~\subref{fig:parameter-behavior-timestep}: each metric is normalized to its own best timestep; recall, particularly for small objects, degrades earlier than mAP. Dashed lines mark the selected $p=1.6$ and $t=150$. All metrics use at most 1000 detections.}
    \label{fig:parameter-behavior}
\end{figure}

In this section, we analyze the impact of \mymethod's design choices on its pseudo-label and zero-shot performance, focusing particularly on hyperparameters of the single-step self-attention extraction, prompt propagation, and segment merging.
The extraction timestep $t$ and the propagation parameter $p$ strongly influence label-generation performance. For extraction, \Cref{fig:parameter-behavior-timestep} shows that our approach's recall degrades across timesteps on the SA-1B validation set. This is because earlier time steps focus more on fine texture, which favors our objective of segmenting objects at varying granularities, whereas later steps capture coarse semantics. We select $t = 150$ for our experiments, which maximizes mAP while keeping strong mAR. For propagation, we show different $p$ values in \Cref{fig:parameter-behavior-p}  from linear propagation ($p = 2$) to increasingly edge-preserving smoothing ($p < 2$) on the SA-1B validation set.  We show that non-linear propagation in our method vastly improves mAR compared to linear propagation. While values closer to $p = 1$ improve recall and isolate more objects, they also lower mAP and introduce 
noisy masks. For our experiments, we prioritize mask precision and thus select $p = 1.6$, where mAP peaks.
\begin{table}[htb]
\begin{subtable}[t]{0.485\linewidth}
    \tablestyle{1.5pt}{1.05}
    \scriptsize
    \begin{tabular*}{0.95\textwidth}{@{\extracolsep{\fill}}lcccccc}
        \toprule
        Levels & $\text{AR}_{1000}$& AR$_S$ & AR$_M$ & AR$_L$ & mAP  \\
        \midrule
        3 & 16.3 & 2.1 & 11.0 & 32.4 & 1.9 \\
        5 & 17.7 & 2.5 & 12.4 & 34.6 & 1.8  \\
        \rowcolor{gray!15}
        6 & \textbf{18.1} & \textbf{2.5} & \textbf{12.7} & \textbf{35.0} & 1.8 \\
        \bottomrule
    \end{tabular*}
    \caption{Number of levels in the map merging step}
    \label{tab:merging-cut-set}
\end{subtable}%
\hfill
\begin{subtable}[t]{0.485\linewidth}
    \tablestyle{1.5pt}{1.05}
    \scriptsize
    \begin{tabular*}{0.95\textwidth}{@{\extracolsep{\fill}}lcccccc}
        \toprule
        NMS $\tau_{\mathrm{IoU}}$ & $\text{AR}_{1000}$ & AR$_S$ & AR$_M$ & AR$_L$ & mAP \\
        \midrule
        $0.5$ & 15.4 & 2.1 & 10.8 & 30.1 & 1.8 \\
        $0.7$ & 17.0 & 2.4 & 11.9 & 33.2 & 1.8 \\
        $0.8$ & 17.8 & 2.5 & 12.4 & 34.6 & 1.8 \\
        \rowcolor{gray!15}
        $0.9$ & \textbf{18.5} & \textbf{2.5} & \textbf{12.9} & \textbf{36.1} & 1.8 \\
        \bottomrule
    \end{tabular*}
    \caption{Cross-level deduplication}
    \label{tab:merging-dedup}
\end{subtable}
\caption{\textbf{\mymethod mAR as a function of number of hierarchy levels and NMS threshold for deduplication.} We find that a higher number of hierarchy levels and a conservative $\tau_{\mathrm{IoU}}$ in our mask merging and deduplication consistently boosts recall. }
\label{tab:label_merging}
\end{table}
\paragraph{Mask Merging} We next ablate the parameters of our post-processing step. Increasing the number of hierarchy levels consistently improves recall, since additional levels help to isolate a bigger variety of granularities. Going from 3 to 6 levels improves mAR from $16.3$ to $18.1$ (see \Cref{tab:merging-cut-set}). To deduplicate across levels, we use a simple area-sorted non-maximum suppression (NMS). We analyze different thresholds in \Cref{tab:merging-dedup} and find that a conservative threshold of $\tau_{\text{IoU}}=0.9$ best retains mask candidates across granularities. Deduplication using a smaller overlap threshold, such as $\tau_{\text{IoU}}=0.5$, reduces recall significantly by 3.1 p.p.\ in mAR. We therefore keep 6 levels and $\tau_{\text{IoU}}=0.9$ for our experiments.

\begin{table}[t]
\begin{subtable}[t]{0.485\linewidth}
    \vspace{0pt}
    \centering
    \tablestyle{1.8pt}{1.05}
    \resizebox{\linewidth}{!}{%
    \begin{tabular}{lcccccc}
        \toprule
        \multirow{2}{*}{\textbf{Configuration}} &
        \multicolumn{2}{c}{\textbf{Things}} &
        \multicolumn{3}{c}{\textbf{Stuff + Things}} &
        \multirow{2}{*}{\textbf{Parts}} \\
        \cmidrule(lr){2-3} \cmidrule(lr){4-6}
        & COCO & LVIS & ADE20K & EntitySeg & SA-1B & PACO \\
        \midrule
        \mymethod labels &
        35.6 & 35.0 & 36.0 & 38.2 & 39.8 & 25.8 \\
        \quad + Crop labels &
        37.4 & 36.9 & 37.6 & 39.7 & 43.5 & 27.1 \\
        \rowcolor{gray!10}
        \quad + CutLER mixing &
       41.0 & 38.7 & 39.8 & 42.2 & 45.5 & 28.9 \\
        \bottomrule
    \end{tabular}%
    }
    \caption{\textbf{Impact of the training supervision on mAR.} Starting from global \mymethod pseudo-labels, the rows add crop-refined child masks and CutLER \mbox{masks, respectively.}}
    \label{tab:training-label-impact}
\end{subtable}%
\hfill
\begin{subtable}[t]{0.485\linewidth}
    \vspace{0pt}
    \centering
    \tablestyle{1.4pt}{1.05}
    \scriptsize
    \resizebox{\linewidth}{!}{%
    \begin{tabular}{llrrcr}
        \toprule
        \multirow{2}{*}{\textbf{Supervision}} &
        \multirow{2}{*}{\textbf{Stage}} &
        \multirow{2}{*}{\textbf{\#\;Images}} &
        \multicolumn{3}{c}{$\textbf{AR}_{1000}$} \\
        \cmidrule(lr){4-6}
        &&& Things & Stuff+Things & Parts \\
        \midrule
        \multirow{2}{*}{CutLER mixing} 
        & Stage 1 & 0.1M & 37.9 & 40.3 & 27.0 \\
        & Stage 2 & 0.1M & 38.7 & 40.7 & 27.1 \\
        \midrule
        + Crop labels  
        & Stage 2 & 0.1M & 39.9 & 42.5 & 28.9 \\
        \bottomrule
    \end{tabular}%
    }
    \caption{\textbf{Self-training strategies.} Self-training with our presented strategies consistently improves results. Self-training was done on 0.1M images.}
    \label{tab:training-stage-impact}
\end{subtable}
\caption{\textbf{Impact of label sources and training stages.} (a) Starting from global \mymethod pseudo-labels, the rows add crop-refined child masks and CutLER masks, respectively. Crop refinement adds locally re-inferred, parent-contained child masks. All rows use a $0.1$M image budget. (b) Self-training with our presented strategies significantly improves the model performance by correcting and filling gaps of the initial pseudo-labels. }
\label{tab:training-config-impact}
\end{table}
\paragraph{Training Supervision} Finally, we study the different label sources and training stages for our \mymethod training. Training only on the diffusion-based labels significantly outperforms the competitor SOHES, which likewise does not use detector supervision. Our crop-refined masks from the second round improve fine structures (see \Cref{tab:training-label-impact}), with gains of +3.7 p.p.\ on SA-1B and +1.3 p.p.\ on PACO. CutLER mixing, in contrast, improves performance most strongly on things datasets (+3.6 p.p.\ on COCO). Moreover, stuff+things datasets benefit from coarse-instance coverage. On the EntitySeg dataset, for instance, augmenting with \cutler improves $\text{AR}_{1000}$ by 2.5 p.p. The stage analysis in \Cref{tab:training-stage-impact} reveals that our initial model struggles with visual details because its pseudo-labels are generated from a low-resolution latent. Following the UnSAM configuration for a second round of self-training that merges only global predictions into the initial pseudo-label dataset yields only marginal gains in our setup, improving stuff+things mAR from 40.3 to 40.7. In contrast, our crop-based labels effectively mitigate the lack of details in our second round of fine-tuning and boost performance on the stuff+things, things, and parts datasets by +2.2 p.p., +2.0 p.p., and \mbox{+1.9 p.p., respectively}.

\section{Conclusion}
We proposed \mymethod, the first method to extract dense multi-granular instance pseudo-labels from pre-trained diffusion models without retraining or supervision. Our extensive experiments show not only the generality of pseudo-label generation but also the effectiveness of training segmentation models unsupervised using our pseudo-labels for strong zero-shot performance. In addition, we show that our models can serve as an effective initialization for semi-supervised learning, even surpassing a fully supervised reference with significantly fewer labeled images. This suggests that pre-trained diffusion models learn fine-grained scene representations. As a limitation, our pseudo-labels struggle with particularly small objects that cannot be recognized at the latent resolution.
Additionally, the diffusion representation may encode not only semantics but also texture and therefore sometimes oversegment larger segments in the image, such as the sky. Adopting coarse-to-fine approaches and objectness ranking, as in the SSL-based literature, can be practical solutions to address these issues. Finally, our propagation framework is flexible in prompts and in-model propagation, enabling future extensions such as class-conditioned cross-attention prompting or affinity extraction from generative video models to segment objects over time.

\bibliography{main}
\bibliographystyle{tmlr}

\appendix

\section{Appendix}

\subsection{Pseudo-Label Generation}
\label{sec:app_nonlocal}
For our pseudo-label generation, we use an efficient solver following iterative Gauss-Jacobi introduced in \citep{Elmoataz08} which is described in \Cref{alg:non-linearprop}. We use a vectorized implementation and propagate all prompts in parallel.
\begin{algorithm}[tp!]
\caption{Non-linear prompt propagation: A grid of one-hot prompts is propagated into soft object maps }\label{alg:pseudolabel}
\begin{algorithmic}[1]
\Require image $I$; SD UNet; layer weights $\{w_l\}$; timestep $t$;
         anchor $\lambda$; power $p$; threshold $\tau_{\text{prop}}$
\Ensure  soft object maps $\{f^{(1)},\dots,f^{(K)}\}$
\State \textbf{// Self-attention extraction}
\State $z \gets \mathrm{VAE}(I)$
\State $ \epsilon_{t}  \gets \mathrm{UNet}(z, t)$ \Comment{Denoise at timestep t}
\State $\mathbf{A} \gets \sum_{l\in L} w_l \, \tfrac{1}{n_h}\sum_{h\in n_{h}}\mathbf{A}_{(l,h)}$
       \Comment{Collect $N\times N$ affinity}
\State \textbf{// Prompt propagation}
\State $\{f_0^{(1)},\dots,f_0^{(K)}\} \gets$ equidistant one-hot grid prompts on the latent
\For{$k = 1$ \textbf{to} $K$}
    \State $\mathbf{f} \gets \mathbf{f}^{(0)}_{k}$
    \Repeat
        \State $g_i \gets \sqrt{\sum_j A_{ij}(f_j - f_i)^2}\quad \forall i$
        \State $\gamma_{ij} \gets A_{ij}\big(g_i^{\,p-2} + g_j^{\,p-2}\big)\quad \forall i,j$
        \State $f_i \gets \dfrac{\lambda\, f_{0,i}^{(k)} + \sum_j \gamma_{ij} f_j}
                            {\lambda + \sum_j \gamma_{ij}}\quad \forall i$
     \Until{ $ \|\mathbf{f}^{t+1} - \mathbf{f}^{t}\|_2^2  \leq \tau_{\text{prop}} $}
    \State $\mathbf{f}^{(k)} \gets \mathbf{f}$
\EndFor
\State \Return $\{f^{(1)},\dots,f^{(K)}\}$
\end{algorithmic}
\label{alg:non-linearprop}
\end{algorithm}
After producing our soft maps, we need to generate instance mask proposals from them. Our next steps consist of map merging and post-processing. Here, we show these two steps in detail \mbox{in \Cref{alg:maskmerge}}.
\begin{algorithm}[H]
\caption{Mask merging: Taking the initial propagated soft object maps as input, the mask merging step produces a final set of instance mask candidates}\label{alg:maskmerge}
\begin{algorithmic}[1]
\Require soft object maps $\{\mathbf{f}_{(1)},\dots,\mathbf{f}_{(K)}\}$; thresholds $\mathcal{H}=\{h_1,\dots,h_L\}$;
         min.\ area $a_{\min}$; IoU thr.\ $\tau$; cap $M_{\max}$
\Ensure  pseudo-mask set $\mathcal{M}$
\State \textbf{// Mask merging}
\State  $\textbf{p}_{k} = \textbf{f}_{k}/{\sum_i f_{k,i}} \quad \forall k$
       \Comment{Normalize to distributions}
\State   $d_\mathrm{KL}  \gets \frac{1}{2} \biggl( D_\mathrm{KL}\Bigl(\mathbf{p}_{(k)} \!\parallel\! \mathbf{p}_{(k')}\Bigr) + D_\mathrm{KL}\Bigl(\mathbf{p}_{(k')} \!\parallel\! \mathbf{p}_{(k)}\Bigr) \biggr)\ $ \Comment{Compute pairwise distance between prompts}

\State $\mathcal{P} \gets \emptyset$
\For{$h \in \mathcal{H}$}
    \State $\{C_1^h,\dots,C_{K_h}^h\} \gets \textsc{AgglomCluster}(\mathbf{D}, h)$
    \State $\bar{\mathbf{p}}_c^{\,h} \gets \tfrac{1}{|C_c^h|}\sum_{k\in C_c^h} \mathbf{p}_{(k)}\quad \forall c$
    \State $\tilde{\mathbf{p}}_c^{\,h} \gets $\textsc{Upsample} $(\bar{\mathbf{p}}_c^{\,h},\, H\!\times\!W)\quad \forall c$
    \State $M^h(x) \gets \arg\max_c \tilde{\mathbf{p}}_c^{\,h}(x)$
    \For{each label $c$ in $M^h$}
        \State $r \gets \textsc{ConnectedComponents}(M^h)$
        \If{$\mathrm{area}(r) \ge a_{\min}$}
        \State $\mathcal{P} \gets \mathcal{P} \cup \{r\}$
        \EndIf
        \EndFor
    \EndFor
\State \textbf{// Area-descending non-maximum suppression}
\State sort $\mathcal{P}$ by descending area; $\mathcal{M} \gets \emptyset$
\For{$p \in \mathcal{P}$}
    \If{$|\mathcal{M}| < N_{\max}$ \textbf{and} $\max_{q\in\mathcal{M}} \mathrm{IoU}(p,q) \le \tau_{\text{IoU}}$}
        \State $\mathcal{M} \gets \mathcal{M} \cup \{p\}$
    \EndIf
\EndFor
\State \Return $\mathcal{M}$
\end{algorithmic}
\end{algorithm}

\def\tabWholeImageInferenceAP#1{
    \begin{table}[#1]
        \tablestyle{2.2pt}{1.0}
        \begin{center}
        \vspace{-8pt}
        \resizebox{\linewidth}{!}{%
        \begin{tabular}{Hllcccccccccccc}
            \multirow{2}{*}{Supervision} & \multirow{2}{*}{Methods} &
            \multirow{2}{2.2cm}{\centering Backbone\\\rule{0pt}{8pt}(\# params)} & \multirow{2}{1.0cm}{\centering \# images} && \multirow{2}{*}{Avg.} &
            \multicolumn{6}{c}{Datasets with Whole Entities} && \multicolumn{2}{c}{Datasets w/ Parts} \\
            \cline{7-12} \cline{14-15}
            &&&&&& COCO & LVIS & ADE & Entity & SA-1B & COCONut && PtIn & PACO \\

            \Xhline{0.8pt}
            \multirow{3}{*}{Self-sup.} & \ours & RN-50 (23M) & 0.4M &&  -- & 3.5 &  3.2 &  4.0 &  4.4 &  10.8 & -- &&  3.4 &  1.2 \\
            \rowcolor{gray!10}
            & \ours & RN-50 (23M) & 0.01M && 2.3 & \bf 1.6 & \bf 1.7 & 1.9 & 2.7 & 6.0 & 1.4 && \bf 1.8 & \bf 0.9 \\
            \rowcolor{gray!10}
            & \mymethod (GranSamp) & RN-50 (23M) & 0.01M && \bf 3.0 & 1.5 & 1.6 & \bf 2.3 & \bf 5.2 & \bf 9.8 & \bf 2.0 && 1.2 & 0.7 \\
            & \mymethod (mix c+p) & RN-50 (23M) & 0.01M && \bf  & 1.6 & 2.4 & 3.1 &  4.8 &  10.2 &  1.4 && -- & 0.8 \\
            \Xhline{0.8pt}
            & stage 2 &  &  &&  &  &  &  &   &   &   &&  &  \\
           & \ours & RN-50 (23M) & 0.01M &&  & 2.6 & 2.6 & 2.5 & 3.0 & 6.2 & 2.2 && 2.6 &  1.0 \\

        \end{tabular}}%
        \end{center}\vspace{-1pt}
        \caption{
        Zero-shot evaluation of pre-trained models on unsupervised image segmentation benchmarks.
        The evaluation metric is mean average precision (mAP).
        }
        \label{tab:whole-image-ap}
    \end{table}
}

\def\tabWholeImageInferenceAPiou#1{
    \begin{table}[#1]
        \tablestyle{2.2pt}{1.0}
        \begin{center}
        \vspace{-8pt}
        \resizebox{\linewidth}{!}{%
        \begin{tabular}{Hllcccccccccccc}
            \multirow{2}{*}{Supervision} & \multirow{2}{*}{Methods} &
            \multirow{2}{2.2cm}{\centering Backbone\\\rule{0pt}{8pt}(\# params)} & \multirow{2}{1.0cm}{\centering \# images} && \multirow{2}{*}{Avg.} &
            \multicolumn{6}{c}{Datasets with Whole Entities} && \multicolumn{2}{c}{Datasets w/ Parts} \\
            \cline{7-12} \cline{14-15}
            &&&&&& COCO & LVIS & ADE & Entity & SA-1B & COCONut && PtIn & PACO \\

            \Xhline{0.8pt}
            \multirow{3}{*}{Self-sup.} & \ours & RN-50 (23M) & 0.4M &&  -- & -- &  -- &  -- &  -- &  -- & -- && -- &  -- \\
            \rowcolor{gray!10}
            & \ours & RN-50 (23M) & 0.01M && 4.5 & \bf 3.6 &  3.4 & 4.3 & 4.9 &  11.0 & 3.0 && \bf 4.2 & \bf 1.8 \\
            \rowcolor{gray!10}
            & \mymethod (GranSamp) & RN-50 (23M) & 0.01M && \bf 5.8 & 3.4 & 3.0 &  5.0 & \bf 9.4 &  17.2 & \bf 3.7 && 3.0 & 1.5 \\
            & \mymethod (mix c+p) & RN-50 (23M) & 0.01M && \bf  & 3.3 & \bf 3.8 & \bf 6.0 & 8.7 & \bf 17.7 &  3.1 &&  & 1.6 \\

        \end{tabular}}%
        \end{center}\vspace{-1pt}
        \caption{
        Zero-shot evaluation of pre-trained models on unsupervised image segmentation benchmarks.
        The evaluation metric is mean average precision at iou=0.5 ($\textrm{mAP}_{50}$).
        }
        \label{tab:whole-image-ap-50}
    \end{table}
}

\def\tabWholeImageInferenceAPnew#1{
    \begin{table}[#1]
        \tablestyle{2.2pt}{1.0}
        \begin{center}
        \vspace{-8pt}
        \resizebox{\linewidth}{!}{%
        \begin{tabular}{Hllcccccccccc}
            \multirow{2}{*}{Supervision} & \multirow{2}{*}{Methods} &
            \multirow{2}{2.2cm}{\centering Backbone\\\rule{0pt}{8pt}(\# params)} &
            \multirow{2}{1.0cm}{\centering \# images} &
            \multirow{2}{*}{Avg.} &
            \multicolumn{6}{c}{Datasets with Whole Entities} &
            \multicolumn{1}{c}{Datasets w/ Parts} \\
            \cline{6-11} \cline{12-12}

            & & & & &
            COCO & LVIS & ADE & Entity & SA-1B & COCONut &
            PACO \\

            \Xhline{0.8pt}
            \multirow{3}{*}{Self-sup.} & \ours & RN-50 (23M) & 0.4M &
            -- & 3.5 & 3.2 & 4.0 & 4.4 & 10.8 & -- &
            1.2 \\

            \rowcolor{gray!10}
            & \ours & RN-50 (23M) & 0.01M &
            2.3 & \bf 1.6 & 1.7 & 1.9 & 2.7 & 6.0 & 1.4 &
            \bf 0.9 \\

            \rowcolor{gray!10}
            & \mymethod (GranSamp) & RN-50 (23M) & 0.01M &
             3.3 & 1.5 & 1.6 &  2.3 & \bf 5.2 & 9.8 & \bf 2.0 &
            0.7 \\

            & \mymethod (mix c+p) & RN-50 (23M) & 0.01M & \bf 3.5
            & \bf 1.6 & \bf 2.4 & \bf 3.1 & 4.8 & \bf 10.2 & 1.4 &
            0.8 \\

        \end{tabular}}%
        \end{center}\vspace{-1pt}
        \caption{
        Zero-shot evaluation of pre-trained models on unsupervised image segmentation benchmarks.
        The evaluation metric is mean average precision (mAP).
        }
        \label{tab:whole-image-ap}
    \end{table}
}

\def\tabWholeImageInferenceARnew#1{
    \begin{table}[#1]
        \tablestyle{2.2pt}{1.0}
        \begin{center}
        \vspace{-8pt}
        \resizebox{\linewidth}{!}{%
        \begin{tabular}{Hllcccccccccc}
            \multirow{2}{*}{Supervision} & \multirow{2}{*}{Methods} &
            \multirow{2}{2.2cm}{\centering Backbone\\\rule{0pt}{8pt}(\# params)} &
            \multirow{2}{1.0cm}{\centering \# images} &
            \multirow{2}{*}{Avg.} &
            \multicolumn{6}{c}{Datasets with Whole Entities} &
            \multicolumn{1}{c}{Datasets w/ Parts} \\
            \cline{6-11} \cline{12-12}

            & & & & &
            COCO & LVIS & ADE & Entity & SA-1B & COCONut &
            PACO \\

            \Xhline{0.8pt}
            \multirow{3}{*}{Self-sup.} 

            \rowcolor{gray!10}
            & \ours & RN-50 (23M) & 0.01M &
            30.5 &  35.6 &  32.4 & 32.6 & 31.4 & 28.6 & 29.3 &
            23.6 \\

            \rowcolor{gray!10}
            & \mymethod (GranSamp) & RN-50 (23M) & 0.01M & 31.5 & 34.2 &  31.8 & 33.6 & 35.4 & 34.0 & 28.6 & 
            23.1 \\

            & \mymethod (no CutLER) & RN-50 (23M) & 0.1 M & 
            & 35.6 & 35.0 & 36.0 & 38.2 & 39.8 & 30.3 & 25.8
              \\
            & \mymethod (mix c+p) & RN-50 (23M) & 0.01M &  31.7
            & 34.2 & 32.0  &  33.8 &  35.4 &  34.4 &  28.8 &
            23.3  \\
            & \mymethod (mix c+p) & RN-50 (23M) & 0.1 M & 36.2
            & 38.7  & 36.2 & 37.9 &  40.1 & 40.9 & 32.8 & 26.5
              \\
            & \mymethod & RN-50 (23M) & 0.4M &
            -- & 41.4 & 39.0 & \bf 40.0 &  \bf 42.7 & \bf 45.4 & 35.1 &
            28.9 \\
            \Xhline{0.8pt}
             & stage 2  & & & 
            &  &  &  & &  &  &
              \\
            & \ours & RN-50 (23M) & 0.01M &
             &  35.4 & 32.5 & 32.7 & 31.8 & 29.1 & 29.4 &
            23.8 \\
            & 
            \mymethod & RN-50 (23M) & 0.1M &
            -- & 39.0 & 36.7 & 38.1 * & 40.5 & 42.3 & 33.3 &
            27.0 \\
             & \mymethod (mix c+p) & RN-50 (23M) & 0.1M &
            -- & 39.0 & 37.1 & 38.6 * & \textbf{41.1} & \textbf{42.4} &  & \\
            & 
            \ours & RN-50 (23M) & 0.1M &
            -- & 40.5 & 37.7 & 35.7 & 39.6 & 41.9 & -- &
            27.5 \\
            & 
            \ours & RN-50 (23M) & 0.4M &
            -- & 42.0 & 40.5 & 39.1 & 41.1 & 44.8 & -- &
            29.7 \\
            
        \end{tabular}}%
        \end{center}\vspace{-1pt}
        \caption{
        Zero-shot evaluation of pre-trained models on unsupervised image segmentation benchmarks.
        The evaluation metric is mean average recall (mAR).
        }
        \label{tab:whole-image-ap}
    \end{table}
}

\def\tabWholeImageInferenceAPnewiou#1{
    \begin{table}[#1]
        \tablestyle{2.2pt}{1.0}
        \begin{center}
        \vspace{-8pt}
        \resizebox{\linewidth}{!}{%
        \begin{tabular}{Hllcccccccccc}
            \multirow{2}{*}{Supervision} & \multirow{2}{*}{Methods} &
            \multirow{2}{2.2cm}{\centering Backbone\\\rule{0pt}{8pt}(\# params)} &
            \multirow{2}{1.0cm}{\centering \# images} &
            \multirow{2}{*}{Avg.} &
            \multicolumn{6}{c}{Datasets with Whole Entities} &
            \multicolumn{1}{c}{Datasets w/ Parts} \\
            \cline{6-11} \cline{12-12}

            & & & & &
            COCO & LVIS & ADE & Entity & SA-1B & COCONut &
            PACO \\

            \Xhline{0.8pt}
            \multirow{3}{*}{Self-sup.} 

            \rowcolor{gray!10}
            & \ours & RN-50 (23M) & 0.01M &
            4.6 & \bf 3.6 & 3.4 & 4.3 & 4.9 & 11.0 & 3.0 &
            \bf 1.8 \\

            \rowcolor{gray!10}
            & \mymethod (GranSamp) & RN-50 (23M) & 0.01M &
             6.2 & 3.4 & 3.0 &  5.0 & \bf 9.4 & 17.2 & \bf 3.7 &
            1.5 \\

            & \mymethod (mix c+p) & RN-50 (23M) & 0.01M & \bf 6.3
            &  3.3 & \bf 3.8 & \bf 6.0 & 8.7 & \bf 17.7 & 3.1 &
            1.6 \\

           & \mymethod (mix c+p) & RN-50 (23M) & 0.1M & 
            &  4.1 &  3.4 &  6.2 & 11.1 & 19.7 & 3.4 &
            1.7 \\
            \Xhline{0.8pt}
             & stage 2  & & & 
            &  &  &  & &  &  &
              \\
            & \ours & RN-50 (23M) & 0.01M &
             &  4.7 & 4.4 & 5.2 & 5.3 & 11.2 & 3.9 &
             2.0 \\
             & \ours & RN-50 (23M) & 0.4M &
            -- & 6.7 & 5.3 & 7.2 & 8.0 & 17.7 & -- &
            2.4 \\
        \end{tabular}}%
        \end{center}\vspace{-1pt}
        \caption{
        Zero-shot evaluation of pre-trained models on unsupervised image segmentation benchmarks.
        The evaluation metric is mean average precision at iou=0.5 (mAP50).
        }
        \label{tab:whole-image-ap-iou-noptin}
    \end{table}
}

\begin{figure}[H]
    \centering
    \includegraphics[width=\linewidth]{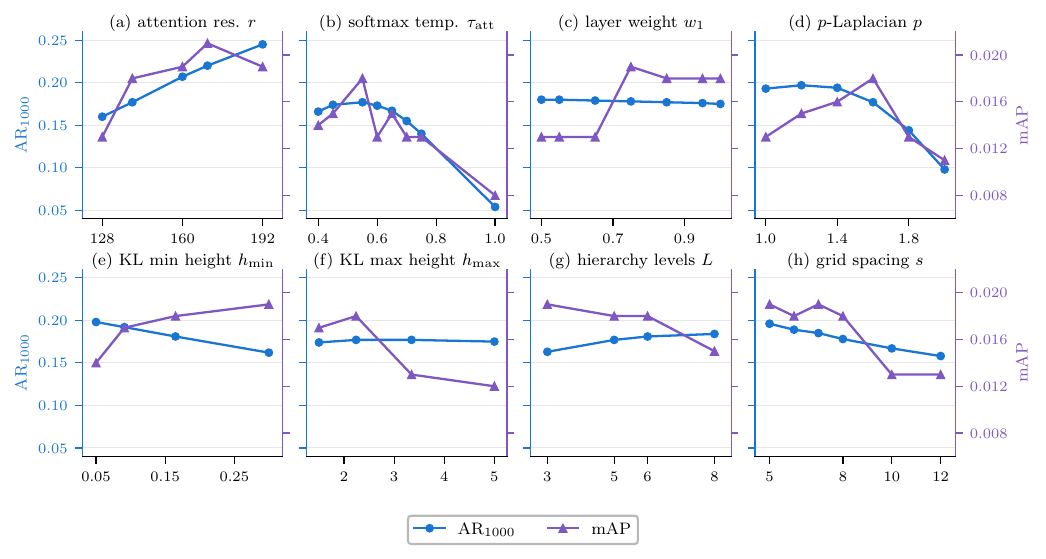}
    \caption{\textbf{Parameter sweeps of \mymethod{} label generation on a SA-1B holdout set.} Each panel varies a single parameter with all others held at a reference configuration.  $\mathrm{AR}_{1000}$ (blue, left axis) and $\mathrm{mAP}$ (purple, right axis) are shown at their true values, using identical scales across all panels to enable direct comparison of parameter sensitivities. In some cases a shared optimum for both metrics (\eg $\tau_{\mathrm{att}}$) can be observed. However, often there is tradeoff, for instance recall being stable but precision decreasing (\eg $w_1$, $L$), and trade-offs of in-domain recall and precision (\eg $p$). Moreover, increasing some parameters incurs higher computational cost, such as grid spacing $s$ and resolution $r$.}
    \label{fig:sa1b-one-factor-sweeps}
\end{figure}
\mymethod{} generates pseudo-labels without training, so we determine all label-generator hyperparameters once on a held-out SA-1B 1k split and keep them frozen across all datasets and downstream training runs. We show the hyperparameter behavior in \Cref{fig:sa1b-one-factor-sweeps}. For our productive configuration, we aim to balance mask precision and recall. In detail, we resize the images to $1120 \times 1120$ $(r=140)$, which is outside the native resolution of SD2 but still yields good segmentation results. We combine features from the top two decoder layers with $w_1 = 0.85$ and $w_2 = 0.15$, which lies on the precision plateau. Affinities are sharpened with $\tau_{\mathrm{att}} = 0.55$, the optimum for recall and precision. Our prompt seeds are placed on an equidistant grid with spacing ($s=6$), retaining 96.6\% of the recall of a denser $s=5$ grid. We set the propagation exponent to $p=1.6$ to prioritize precision (0.018 vs. 0.015 at $p=1.2$, avoiding noisy boundary fragments for downstream student training) with prompt anchoring $\lambda = 1 \times 10^{-5}$. Finally, multi-granularity masks are formed via average-link clustering over $L=6$ log-spaced thresholds between $h_{\min}=0.186$ and $h_{\max}=2.99$, followed by NMS at $\tau_{\mathrm{IoU}} = 0.9$ and noise filtering at $A_{\min} = 100$\,px.

\subsection{CascadePSP Pseudo-Label Post-Processing}
\label{sec:cascadepsp_hyperarameters}
Our post-processing with CascadePSP~\citep{cheng2020cascadepsp} follows UnSAM and discards refined masks with area fraction $> 0.9$ or refinement overlap $\mathrm{IoU}(m, \tilde{m}) < 0.5$. We group mask crops into spatial buckets and pad them to squares with zeros. We execute CascadePSP with FP16 autocast. In UnSAM, the refinement crop side length $L$ scales up to $L_{\max} = 900$. As we generate more large masks that need less refinement, we cap post-processing at $L_{\rm max}=480$ to accelerate refinement. 
\subsection{Training Hyperparameters}
\label{sec:appendix_train_hyperarameters}
For training our segmentation models, we follow UnSAM and use Mask2Former~\citep{cheng2022mask2former} with a ResNet-50 backbone. Moreover, we use three different training dataset budgets: 1\%, 2\%, and 4\%.
All experiments share the same optimizer and architecture: AdamW, learning rate $5 \times 10^{-5}$, weight decay $0.05$, backbone multiplier $0.1$, batch size $16$ on $4$ A100 GPUs, gradient clipping $0.01$, input resolution $1024{\times}1024$, $2{,}000$ queries and loss weights $\lambda_{\mathrm{cls}} = 2$, $\lambda_{\mathrm{mask}} = \lambda_{\mathrm{dice}} = 5$.
The encoder in Round~1 is initialized from DINO ResNet-50, and decoder weights are initialized randomly. Round~2 is initialized from the matching \mbox{Round-1 checkpoint}.
\paragraph{Round~1.}
We train for approximately eight epochs at each budget,  with learning-rate decays at $90\%$ and $96\%$ of the schedule. Warmup is $5{,}000$ iterations. The detector-free model is trained on \mymethod labels only. Both Round-1 configurations with and without detector use hierarchical copy-paste with $p_{\mathrm{cp}}{=}0.5$, small-object paste probability $0.8$, area threshold $4096$ pixels, and paste scale in $[0.5,1.0]$.
\paragraph{Round~2.}
For the label enrichment, we re-infer the Round-1 model on the entire image and the padded parent crops of size $H_{\mathrm{ref}}{\times}W_{\mathrm{ref}}{=}1024{\times}1024$ (with a padding ratio of 0.05).
Parents are the top-$k$ predictions for which the area fraction falls within the interval $[a_{\mathrm{low}},a_{\mathrm{high}}]{=}[0.003,0.30]$ and whose pairwise IoU is no more than 0.30. We retain the predictions that have a score of at least $\theta$ and those children for which the containment is $\tau_{\mathrm{cont}}{=}0.70$. We replace the original masks $ m$ with the new masks $\hat{m}$ when the intersection over union of the two masks exceeds $\tau_{\mathrm{merge}}$, which is set to 0.5.
We set $\theta$ to 0.55, $k$ to 12, and use a cropping NMS of 0.75 (with an intra-crop NMS of 0.40) when CutLER is not used. When augmenting with CutLER, we use $\theta{=}0.65$, $k{=}9$, crop NMS $0.85$ (intra-crop NMS $0.50$), and then append CutLER masks. Independent of the Round-1 training image budget, Round~2 fine-tunes only with $0.1$M crop-refined labels to \mbox{maintain efficiency}.
\paragraph {Semi-Supervised Learning} For semi-supervised fine-tuning, we keep the unsupervised checkpoint as initialization, freeze the ResNet-50 backbone, and fine-tune only the Mask2Former decoder on different image budgets of SAM ground-truth masks. We consider budgets of $100$, $500$, $1$k, $2$k, $5$k, and $10$k labeled images, plus a fully supervised $100$k baseline trained from scratch on the same SAM-GT pool. We compare two initializations at every budget: Our final Round-2 $0.4$M checkpoint and the public UnSAM $0.2$M ($4\%$ SA-1B) checkpoint. The optimizer matches unsupervised training (AdamW, learning rate $5{\times}10^{-5}$, weight decay $0.05$, batch size $16$, copy-paste with $p_{\mathrm{cp}}{=}0.5$).
The schedule consists of $\max(1000,\lfloor B/2\rfloor)$ iterations, including a 10\% warmup phase and then decaying at 90\% and 96\% of the run. We always use the latest checkpoint and assess it on the same zero-shot benchmark as our unsupervised models. 
\newpage
\subsection{Semi-Supervised Results}
\label{sec:results_segmentor}
\begin{figure}[H]
    \centering
    \includegraphics[width=\linewidth]{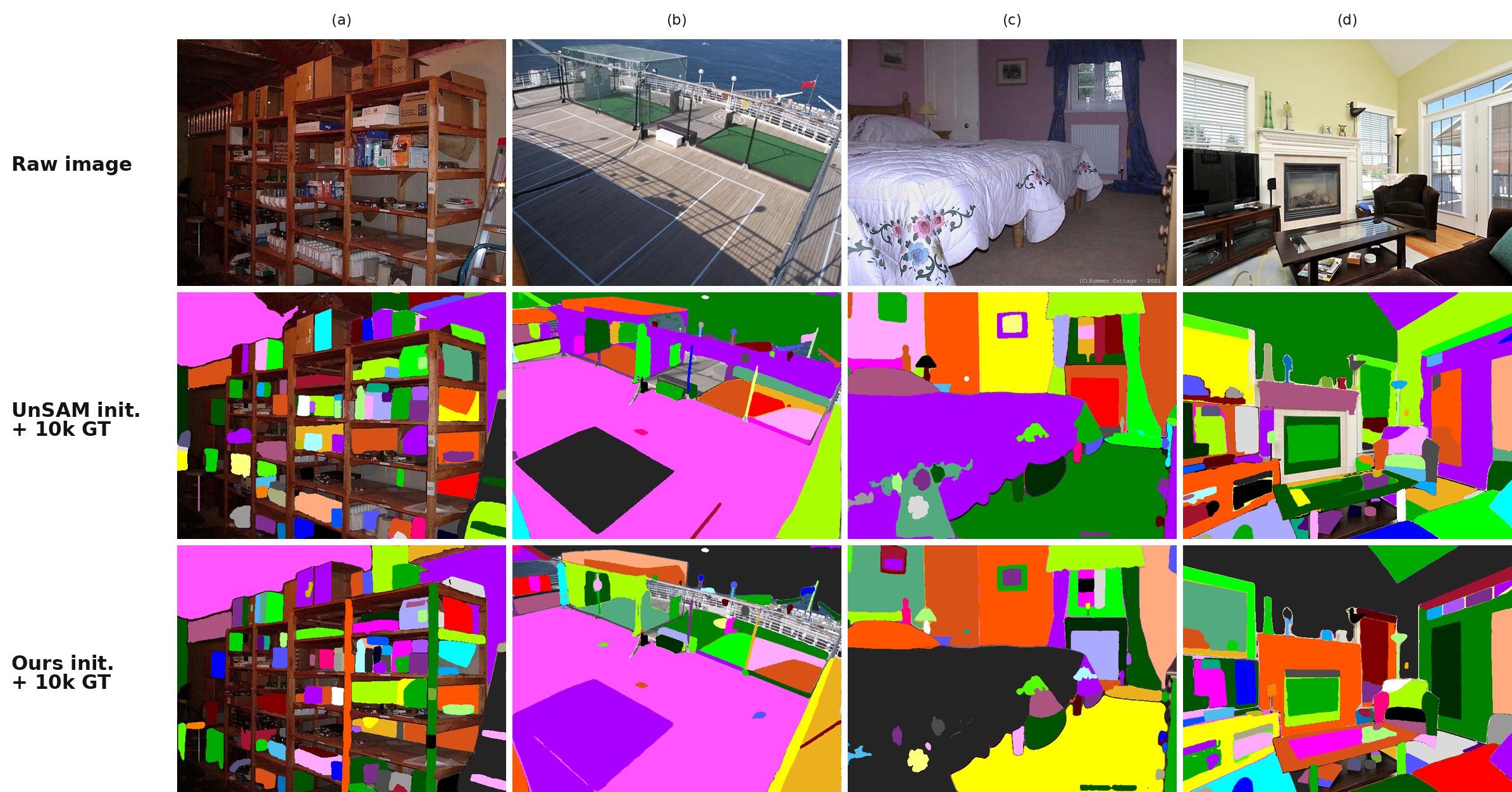}
    \caption{\textbf{Qualitative comparison of semi-supervised instance segmentation on ADE20K.} Each column (a–d) shows a different indoor/outdoor scene. Top: raw input image, middle: predictions of the UnSAM checkpoint fine-tuned on 10k labeled images, bottom: predictions of our \mymethod model fine-tuned on 10k labeled images. Under the same 10k-label budget, initializing from our diffusion-based unsupervised model yields higher mask coverage, crisper object boundaries, and better segmentation of fine structures (\mbox{e.g., furniture} legs, windows, and facade details).}
    \label{fig:placeholder}
\end{figure}

\end{document}